\documentclass[11pt]{article}

\usepackage{acl}

\usepackage{times}
\usepackage{latexsym}

\usepackage[T1]{fontenc}

\usepackage[utf8]{inputenc}

\usepackage{microtype}

\usepackage{inconsolata}

\usepackage{graphicx}

\usepackage{algorithm}
\usepackage{algorithmic}
\usepackage{amsmath}     
\usepackage{amssymb}
\usepackage{mathtools}
\usepackage{amsthm}
\usepackage{tabularx}
\usepackage{booktabs}
\usepackage{multirow}
\usepackage{subcaption}
\usepackage[capitalize,noabbrev]{cleveref}
\usepackage{tcolorbox}

\title{ILRR: Inference-Time Steering Method for Masked Diffusion Language Models}

\author{
 \textbf{Eden Avrahami\textsuperscript{1}},
 \textbf{Eliya Nachmani\textsuperscript{2}},
\\
 \textsuperscript{1}Blavatnik School of Computer Science, Tel Aviv University
 \\
 \textsuperscript{2}School of Electrical and Computer Engineering, Ben-Gurion University of the Negev
\\
\texttt{ea1@mail.tau.ac.il, eliyanac@bgu.ac.il}
}

\begin{document}
\maketitle
\begin{abstract}
Discrete Diffusion Language Models (DLMs) offer a promising non-autoregressive alternative for text generation, yet effective mechanisms for inference-time control remain relatively underexplored. Existing approaches include sampling-level guidance or trajectory optimization mechanisms.
In this work, we study the paradigm of reference-based latent steering for DLMs. We introduce Iterative Latent Representation Refinement (ILRR), an efficient framework for steering DLMs using a reference text as a high-level semantic blueprint. ILRR extracts and injects reference-derived semantic signals into the evolving activations of the generated sequence, enabling tunable transfer of coarse properties such as sentiment. We further introduce Spatially Modulated Steering, an extension that enables long-form generation to be guided by shorter references by regulating intensity across the sequence. Empirically, we demonstrate that ILRR achieves effective control on LLaDA and MDLM architectures with low computational overhead, requiring only one additional parallel forward pass per denoising step. Under comparable compute budgets, ILRR improves attribute accuracy over baselines by 10$\%$ to 60$\%$ points. Our results suggest that the iterative, global denoising process makes DLMs a natural substrate for effective sequence-wide activation-level control.
\end{abstract}

\begin{figure}[t]
    \centering
    \includegraphics[trim=0.5cm 0.7cm 0.7cm 2.7cm, clip, width=\linewidth]{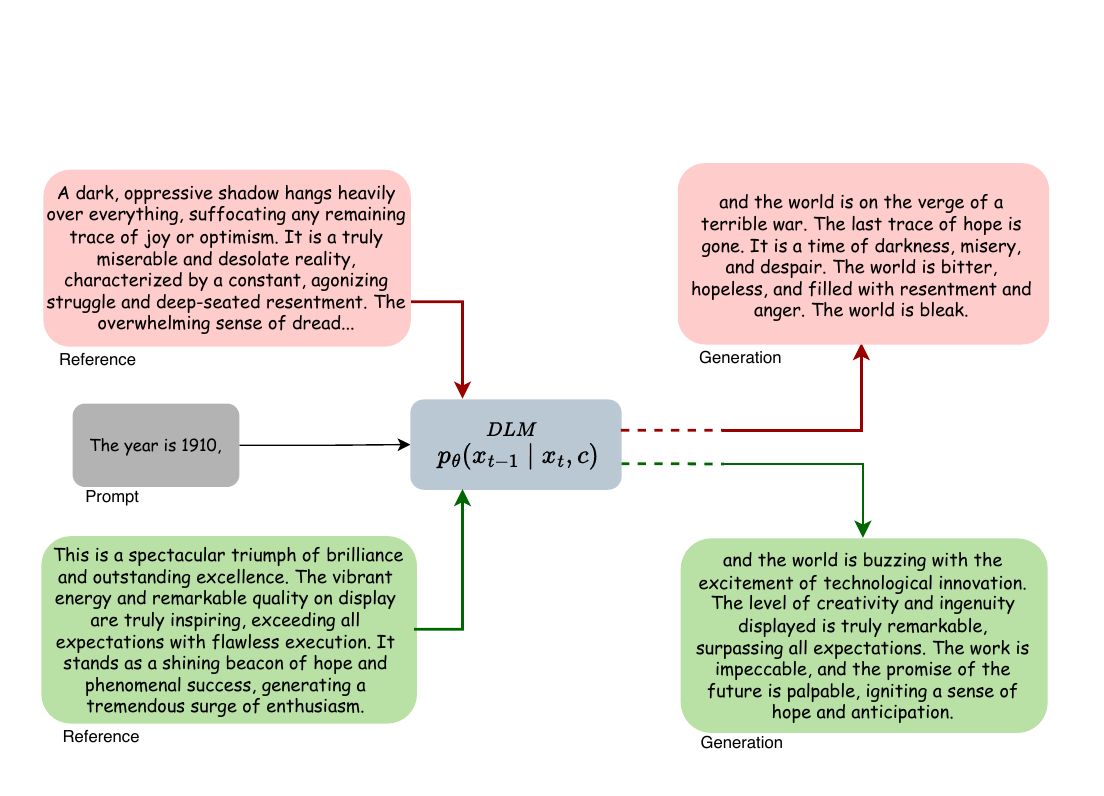}
    \caption{References and corresponding generations for positive and negative sentiment. ILRR guides generations toward the coarse semantics of the reference.}
    \label{fig:ilrr_examples}
\end{figure}

\section{Introduction}
Discrete diffusion language models (DLMs) have recently emerged as a promising non-autoregressive framework for text generation \citep{austin2021structured, sahoo2024simple}. Recent scaled models such as Dream \citep{ye2025dream7bdiffusionlarge} and LLaDA \citep{nie2025largelanguagediffusionmodels} demonstrate strong performance, even on complex tasks such as mathematical reasoning and coding. Unlike autoregressive (AR) models, which generate text sequentially \citep{NIPS2017_3f5ee243, bengio2003neural, radford2018gpt}, DLMs generate through a global iterative procedure, reconstructing tokens by sampling from a learned reverse denoising process.

Despite this progress, inference-time control for DLMs remains relatively underexplored. Inference-time steering is particularly appealing as it allows for flexible modulation of model behaviors without the cost of parameter fine-tuning. Existing DLM steering methods primarily operate at the sampling or trajectory level, including classifier guidance, classifier-free mechanisms, remasking policies and inference-time trajectory optimization 
\citep{singhal2025generalframeworkinferencetimescaling, dang2026inferencetime, schiff2025simple, wang2026remasking, jazbec2025learningunmaskingpoliciesdiffusion}.
While effective, these approaches often require external classifiers, multiple candidate sequences or iterative resampling, limiting their flexibility or increasing inference cost.
A complementary direction is representation-level control, where model behavior is steered by modifying internal activations. In the AR domain, this approach has proven effective, offering practical advantages such as tunable control strength, context efficiency, robustness to prompt brittleness, and stronger enforcement of behaviors weakly expressed under standard prompting or decoding
 \citep{stolfo2025improving, liu2024incontextvectorsmakingcontext, rimsky-etal-2024-steering, wang-etal-2025-beyond-prompt,park2024linearrepresentationhypothesisgeometry}. This area remains relatively less explored for DLMs, which create a unique opportunity by exposing global, sequence-wide representations throughout the iterative denoising process. Alongside this, reference-based guidance has proven effective in continuous image diffusion, where the iterative and global nature of the denoising process enables efficient plug-and-play control using a single reference image for coarse semantic guidance \citep{tumanyan2022plugandplaydiffusionfeaturestextdriven, choi2021ilvrconditioningmethoddenoising}. Together, these advances create a natural opportunity to explore whether reference-based activation guidance can achieve effective control for DLMs.

In this work, we propose \textbf{Iterative Latent Representation Refinement (ILRR)}, an efficient and adaptable framework for latent steering in DLMs. Inspired by the Iterative Latent Variable Refinement (ILVR) technique in continuous image diffusion \citep{choi2021ilvrconditioningmethoddenoising}, ILRR uses a reference text as a high-level semantic blueprint that guides the generation. It iteratively extracts steering signals from a noise-corrupted reference and injects them into selected hidden states of the generated sequence. This nudges the denoising trajectory toward the coarse properties encoded by the reference while allowing the model to determine the concrete wording, syntax and prompt-specific content. As a plug-and-play method, ILRR enables flexible control over semantic attributes without task-specific classifiers or additional training, requiring only one additional parallel forward pass per denoising step.
We evaluate ILRR on standard controlled generation benchmarks, including toxicity and sentiment steering, using LLaDA and MDLM as base models, and show that ILRR consistently outperforms competitive inference-time baselines under comparable compute budgets. We further test ILRR's robustness on detoxification and topic control, both showing effective steering. Our results indicate that reference-based latent steering offers an effective approach for DLMs, leveraging the iterative global nature of the generation process.
\textbf{Our main contributions are as follows:}
(i) We introduce ILRR, a learning-free framework for latent steering in DLMs, combining activation-level control with plug-and-play reference guidance, with tunable parameters controlling steering intensity. (ii) We further propose the spatially modulated steering extension, which allows effective guidance when the reference sequence is significantly shorter than the target output.
(iii) We demonstrate that ILRR achieves strong controllability with low computational overhead, improving steering accuracy by 10\% to 60\% points over competitive baselines.

\section{Related Work}
\label{sec:related_work}
\textbf{Discrete diffusion language models} have emerged as a robust non-AR alternative for text generation, employing architectures that typically utilize iterative denoising or unmasking processes to reconstruct masked sequences \citep{austin2021structured, sahoo2024simple, nie2025largelanguagediffusionmodels, ye2025dream7bdiffusionlarge, lou2024discretediffusionmodelingestimating, shi2024simplified, yu2025discretediffusionlargelanguage, vonrütte2026scalingbehaviordiscretediffusion, shariatian2026latentaugmenteddiscretediffusionmodels, zhou2026next, arriola2025block, vonrütte2025generalizedinterpolatingdiscretediffusion, wang2026remasking, liu2025wedlmreconcilingdiffusionlanguage, kollovieh2026interpolatingdiscretediffusionmodels, he-etal-2023-diffusionbert, gong2025scaling}.

\textbf{Controllable generation} is well-established in AR modeling. Methods include utilization of gradients or auxiliary models, discriminators or decoding-time experts \citep{dathathri2020plugplaylanguagemodels, Yang_2021, krause-etal-2021-gedi-generative, liu-etal-2021-dexperts, lovelace-etal-2024-diffusion}.

\textbf{Activation steering} is another central paradigm, which modifies internal activations to control model's generation, often using steering directions derived from contrastive data or interpretable features. Such methods typically employ an update rule of the form $h \leftarrow h + \alpha v$, modifying the hidden states $h$ during generation using a steering vector $v$ scaled by a factor $\alpha$ \citep{turner2024steeringlanguagemodelsactivation,zou2025representationengineeringtopdownapproach,rimsky-etal-2024-steering,arditi2024refusal,stolfo2025improving,liu2024incontextvectorsmakingcontext,wang-etal-2025-beyond-prompt, park2024linearrepresentationhypothesisgeometry, wang2025semantics, li2024inferencetimeinterventionelicitingtruthful, lee2025programming, peters-etal-2018-deep, sharma2026cold}.

\textbf{Controllable generation methods for DLMs} include reward-guided and particle sampling, search algorithms, classifier-based and classifier-free guidance, learned policies, differentiable constraints or trajectory optimization mechanisms \citep{singhal2025generalframeworkinferencetimescaling, dang2026inferencetime, ye2025implicit, zhang2026inference, schiff2025simple, jazbec2025learningunmaskingpoliciesdiffusion, cardei2026constrained, chen2025optimizingdecodingpathsmasked, tejaswi2026entropyawarerewardguidance, peng2026pathplanningmaskeddiffusion, zhao2026d1,li2024derivativefreeguidancecontinuousdiscrete, phunyaphibarn2026rewardguideddiscretediffusioncleansample, padole2025improvingtextstyletransfer}.

\textbf{Reference-based diffusion guidance}
approaches have demonstrated effectiveness for controllable generation in the image domain, utilizing reference-extracted signals to guide the generation process \citep{choi2021ilvrconditioningmethoddenoising,tumanyan2022plugandplaydiffusionfeaturestextdriven, wang2025controllablereferenceguideddiffusion}.

\section{Background}
\subsection{Diffusion Generative Models}
Diffusion models are a class of generative models that learn to reverse a gradual corruption process. Generation begins with a sample from a high-entropy prior distribution $x_T$ (e.g., pure noise) and iteratively refines it over timesteps $t = T, \dots, 1$ to produce a clean sample $x_0$. Formally, this framework consists of two main processes:
\textbf{(i) Forward process} $q(x_t|x_0)$, which gradually corrupts clean data $x_0$ into noise $x_T$ according to a fixed noise schedule. \textbf{(ii) Reverse process} $p_\theta(x_{t-1}|x_t)$, a model trained to reverse this corruption process, by estimating the clean data or the noise component at each step.
While this structure is shared, the mathematical definition of these processes differs between the continuous and discrete domains.

\subsubsection{Continuous Diffusion Models}
In continuous domains (e.g., images), the forward process $q(x_t|x_0)$ typically adds Gaussian noise to the data. For a clean data point $x_0$, the noisy state $x_t$ is defined as $x_t = \sqrt{\bar{\beta}_t}x_0 + \sqrt{1 - \bar{\beta}_t}\epsilon, \quad \epsilon \sim \mathcal{N}(0, I)$
where $\bar{\beta}_t$ is the noise schedule. The reverse process $p_\theta(x_{t-1}|x_t)$ is a network trained to remove this noise by predicting $\epsilon$ to gradually recover the original data \citep{sohldickstein2015deepunsupervisedlearningusing, ho2020denoising}.

\subsubsection{Discrete Diffusion Language Models}
DLMs \citep{austin2021structured} adapt the framework to the discrete nature of language. We consider a sequence $x_0 = (x_0^1, x_0^2, \dots, x_0^N)$ where each token $x_0^i$ belongs to a vocabulary $\mathcal{V}$, and introduce a special mask token $[M]$.
\\
\textbf{Forward Process.}
The forward process $q$ gradually corrupts the clean sequence $x_0$ by substituting discrete tokens with the mask token $[M]$. The marginal distribution $q(x_t | x_0)$ allows sampling any intermediate state $x_t$ directly:
\begin{equation}
\label{eq:forward_disc}
    q(x_t | x_0) = {Cat}(x_t ; x_0, \bar{\alpha_t})
\end{equation}
where $Cat(\cdot)$ is the categorical distribution over the vocabulary $\mathcal{V}$ with the noise schedule $\bar{\alpha_t}$, and each token $x_t^i$ is either the original $x_0^i$ or $[M]$. As $t \to T$, the sequence becomes fully masked.
\\
\textbf{Reverse Process.}
The reverse process $p_\theta(x_{t-1} | x_t)$ reconstructs the original tokens from the masked tokens. DLMs typically employ a denoising network $p_\theta(x_0 | x_t)$, trained to predict $x_0$ from $x_t$. The reverse transition is then computed using the posterior of the forward process:
\begin{equation}
    p_\theta(x_{t-1} | x_t) \propto q(x_{t-1} | x_t, x_0 \sim p_\theta(x_0 | x_t))
\end{equation}
Practically, the model predicts tokens for the masked positions and then re-samples a partially masked sequence $x_{t-1}$.
\\
\textbf{Network Structure.} The network $p_\theta(x_0 | x_t)$ is typically parameterized as a deep neural network with $D$ layers (blocks), which we denote by $L_{p_\theta} = (l_1, \dots, l_D)$. Given an input sequence $x = (x^1, \dots, x^N)$ of length $N$, let $(h_x^{(1)}, \dots, h_x^{(D)})$ denote the sequence of hidden states (the residual stream) produced by the network, where $
        h_x^{(i)} = (h_{x^1}^{(i)}, \dots, h_{x^N}^{(i)}), \quad h_{x^j}^{(i)} \in \mathbb{R}^d
    $
    such that $h_{x^j}^{(i)}$ represents the hidden vector for token $x^j$ at layer $l_i$, and $d$ is the model’s hidden dimension. Finally, a prediction head $f_{head}(\cdot)$ maps the final hidden states to vocabulary logits: $f_{head}\big(h_x^{(D)}\big) \in \mathbb{R}^{N \times |\mathcal{V}|}$.
    The denoising distribution over the vocabulary is then given by
        $p_\theta(x_0 | x_t) = softmax\big(f_{head}(h_x^{(D)})\big)$.
\\
\textbf{Conditioning Model.} The denoising model may additionally be conditioned on a prefix prompt $c$, in which case we write the denoising network as $p_\theta(x_0 | x_t, c)$ and the corresponding reverse transition as $p_\theta(x_{t-1} | x_t, c)$.

\subsection{Iterative Latent Variable Refinement}
ILVR \citep{choi2021ilvrconditioningmethoddenoising} is a guidance method originally developed for continuous image diffusion. It conditions generation by aligning the low-frequency components of the generated image with those of a reference image $y$.
Let $\phi(\cdot)$ be a linear low-pass filter. In each reverse step $t$, ILVR generates a candidate $x'_{t-1}$ from the model and mixes it with a corrupted version of the reference $y_{t-1} \sim q(y_{t-1}|y)$ (Eq.5 from \citet{choi2021ilvrconditioningmethoddenoising}):
\begin{equation}
    x_{t-1} = \phi(y_{t-1}) + (I - \phi)(x'_{t-1})
\end{equation}
By enforcing consistency in the low-frequency domain defined by $\phi$, ILVR locks the coarse attributes of the generation to the reference $y$, while the high-frequency details are derived from the model's proposal. This allows for diverse generations that remain semantically aligned with the reference.

\begin{figure*}[t]
    \centering
    \includegraphics[trim=0.5cm 0.1cm 0.5cm 0.1cm, clip, width=\textwidth]{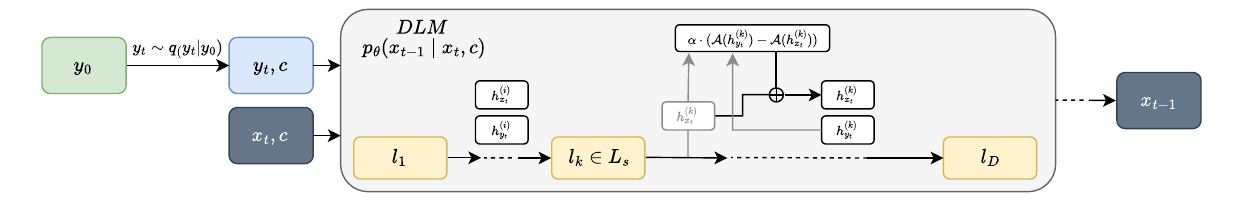}
\caption{Overview of ILRR's steering process within the latent space. The noised reference $y_t$ is processed parallel to $x_t$. Semantic signals are extracted via $\mathcal{A}$ and injected into the generation's hidden states at target layers $L_s$.}
    \label{fig:ilrr_diagram}
\end{figure*}

\section{Method}
Inspired by the principle of ILVR, given a trained DLM $p_\theta(x_{t-1} | x_t, c)$, we aim to generate a text $x_0 \in \mathcal{V}^N$ that shares its high-level semantics with a given reference text $y_0 \in \mathcal{V}^N$. ILRR achieves this by iteratively refining the latent representations of the generation sequence with coarsened semantic signals derived from the reference sequence. Given a total of $T$ denoising steps $1, \dots , T$, refinement is applied in a desired subset of steps $T_s \subseteq \{1, \dots , T \}$, to a targeted subset of hidden states (residual-stream activations) produced from layers $L_s \subseteq L_{p_\theta}$, empirically found to best encode high-level semantic attributes (\Cref{sec:analysis_and_ablation}).

\subsection{Iterative Latent Representation Refinement}
\label{sec:ilrr}
The ILRR procedure at each timestep $t \in T_s$ is defined as follows. First, to ensure that the representations are distributionally aligned, we corrupt the reference sequence $y_0$ to the same corruption level as the current generation state $x_t$ by sampling $y_t \sim q(y_t | y_0)$ (\Cref{eq:forward_disc}). We then perform two parallel forward passes through the denoising model: $p_\theta(\cdot | y_t, c)$ and $p_\theta(x_{t-1} | x_t, c)$. As both sequences propagate through the model, we intervene after each specified target layer $l_k \in L_s$:
Let $h_{x_t}^{(k)}$ and $h_{y_t}^{(k)}$ denote the computed activations for the generation $x_t$ and the noised reference $y_t$, respectively, in layer $l_k$. Given a dimension-preserving semantic-extractor operation $\mathcal{A}(\cdot)$ and a steering scale $\alpha$, we refine the generation's activations via the update rule: 
\begin{equation}
\label{eq:ilrr_rule}
    h_{x_t}^{(k)} \leftarrow h_{x_t}^{(k)} + \alpha \left( \mathcal{A}(h_{y_t}^{(k)}) - \mathcal{A}(h_{x_t}^{(k)}) \right)
\end{equation}
This update rule is grounded in the established activation steering methodology (\Cref{sec:related_work}), with our steering vectors derived dynamically throughout the denoising process.
These modified activations then continue to propagate through the network, shifting all subsequent hidden states and ultimately biasing the prediction of the next state $x_{t-1}$.
Importantly, while the forward pass on $x_t$ is used to sample the next state $x_{t-1}$, the forward pass on $y_t$ is performed solely to extract its hidden representations by the model, while the reference $y_0$ remains static and is re-corrupted at each step. By applying this process iteratively across the denoising steps, ILRR gradually directs the latent representations of the generation toward those of the reference in the high-level semantic space
defined by $\mathcal{A}$, while the scale $\alpha$ controls the strength of this guidance. The complete inference procedure is formalized in \Cref{alg:ilrr}, with the method diagram illustrated in \Cref{fig:ilrr_diagram} and generation examples shown in \Cref{fig:ilrr_examples}.

\begin{algorithm}[H]
\caption{Iterative Latent Representation Refinement}
\label{alg:ilrr}
\begin{algorithmic}[1]
\REQUIRE prefix prompt $c$, trained DLM $p_\theta(x_{t-1} | x_t, c)$, timesteps $T$, reference text $y_0 \in \mathcal{V}^N$, steering scale $\alpha$, target steering layers $L_s \subseteq L_{p_\theta}$, steering steps $T_s$
\ENSURE generated text $x_0 \in \mathcal{V}^N$
\STATE $x_T \leftarrow [M]^N$ \hfill $\triangleright$ Initialize fully masked
\FOR{$t = T, \dots, 1$}
    \STATE $y_t \sim q(y_t | y_0)$
    
    \FOR{k from 1 to $D$}
        \STATE compute $h_{x_t}^{(k)}$ from $h_{x_t}^{(k-1)}$
        \STATE compute $h_{y_t}^{(k)}$ from $h_{y_t}^{(k-1)}$
        \IF{$l_k \in L_s$ \textbf{and} $t \in T_s$}
            \STATE $h_{x_t}^{(k)} \leftarrow h_{x_t}^{(k)} + \alpha \cdot (\mathcal{A}(h_{y_t}^{(k)}) - \mathcal{A}(h_{x_t}^{(k)}))$
        \ENDIF
    \ENDFOR
\STATE compute $x_0$ from ${softmax}(f_{{head}}(h_{x_t}^{(D)}))$  
\STATE compute $x_{t-1} \sim q(x_{t-1} | x_t, x_0)$
\ENDFOR
\STATE Return $x_0$
\end{algorithmic}
\end{algorithm}

\subsection{Extracting High-Level Semantics}
Building upon the understanding that internal representations in language models capture contextualized, interpreted semantic information \citep{peters-etal-2018-deep, tenney-etal-2019-bert, park2024linearrepresentationhypothesisgeometry, zou2025representationengineeringtopdownapproach, turner2024steeringlanguagemodelsactivation}, we adopt 1D average pooling as the semantic-extractor operation $\mathcal{A}$. Our motivation is that, if a semantic attribute is encoded as a shared component in the hidden space, aggregating representations across the sequence should reinforce this common component while attenuating position-specific and lexical variation. We apply this operation as a sliding window along the sequence dimension. Formally, for a sequence of  activation vectors $h_x^{(l)} =(h_{x_1}^{(l)}, \dots, h_{x_N}^{(l)})$, and a kernel size $k$, we compute the pooled representation at position $i$ by averaging the activations within a local window:
\begin{equation}
\mathcal{A}(h_x^{(l)})_i = a_i^{(l)} = \frac{1}{k} \sum_{j=i - \lfloor k/2 \rfloor}^{i + \lfloor k/2 \rfloor} h_{x_j}^{(l)}
\end{equation} 
The resulting sequence $(a_1^{(l)}, \dots, a_N^{(l)})$ aggregates adjacent token activations into a smoothened regional semantic direction. By computing these pooled activations for both the reference and the generation, we can align their underlying semantic directions.

\subsection{Spatially Modulated Steering}
\label{sec:longer_seq}
In certain domains, a reasonable use case is to use a short reference to steer the generation of a longer sequence
($N_{x_0} > N_{y_0}$).
Extending our framework to this setting requires adjusting the mechanism in two key aspects:
(i) Dimensionality Alignment, to enable the computation of valid steering directions between sequences of different lengths; and (ii) Steering Adaptability, to handle the challenge of distributing limited reference signals across a longer sequence in a way that is not overly aggressive (avoiding excessive repetition of identical steering directions), while remaining effective (preserving semantic information without dilution). To address this, we propose \textbf{Spatially Modulated Steering}, an extension that overcomes these challenges through the following mechanisms. \\
\textbf{Interpolated Steering.} Unlike the standard procedure, we operate on sequences of differing lengths. After each layer $l_k \in L_s$, we first apply the semantic extractor $\mathcal{A}$ to the hidden states of both sequences to capture their respective semantic signals. To align their dimensions for the steering update, we downsample the generation’s pooled activations using adaptive average pooling, which compresses the longer sequence into a fixed target length equal to that of the reference, $N_{y_0}$. This operation partitions the generation sequence into uniformly spaced intervals and computes the mean of the hidden states within each interval, such that the $i$-th pooled vector represents the average hidden state over the $i$-th partition. This yields a length-aligned representation while preserving dominant semantic directions through aggregation along the sequence dimension. The steering directions are then formed by upsampling the difference between these aligned representations back to length $N_{x_0}$ via linear interpolation. This interpolation estimates higher-resolution values by taking weighted averages of neighboring positions, effectively spreading the reference’s semantic signal smoothly across the longer generation window. Formally, let $\mathcal{I}_{\downarrow}: \mathbb{R}^{N_{x_0} \times d} \to \mathbb{R}^{N_{y_0} \times d}$ denote the adaptive pooling operation and $\mathcal{I}_{\uparrow}: \mathbb{R}^{N_{y_0} \times d} \to \mathbb{R}^{N_{x_0} \times d}$ denote the linear interpolation operation. The spatially aligned steering directions $\hat{\Delta}^{(k)}$ are computed as:
\begin{equation}\label{eq:down_up_sample}
    \hat{\Delta}^{(k)} = \mathcal{I}_{\uparrow} \left( \mathcal{A}(h_{y_t}^{(k)}) - \mathcal{I}_{\downarrow}(\mathcal{A}(h_{x_t}^{(k)})) \right)
\end{equation}
This operation effectively stretches the dense semantic blueprint of the short reference to cover the longer generation window, ensuring that the guiding signal is continuously distributed.
\\    
\textbf{Modulated Steering Intensity.}  
To prevent dense reference signals from being excessively injected across a broader sequence, we modulate the steering strength spatially along the generation. Specifically, we compute a wave-shaped modulation vector $\mathbf{w} \in \mathbb{R}^{N_{x_0}}$ that smoothly varies the steering intensity across positions. The waveform vector (e.g., cosine) is scaled such that its peaks apply the full steering strength $\alpha$, while its troughs apply minimal steering.
Formally, employing a cosine wave pattern with frequency $f$, the modulation weight $\mathbf{w}_i$ at position $i$ is defined as:
\begin{equation}\label{eq:my_wave}
    \mathbf{w}_i = \frac{\alpha}{2} \left[ 1 + \cos \left( \frac{2\pi f \cdot i}{N_{x_0}} \right) \right]
\end{equation}
where the term $\frac{i}{N_{x_0}}$ spatially normalizes the position, ensuring that the wave pattern is distributed proportionally across the sequence regardless of length, while the outer operations scale the output to the range $[0, \alpha]$.
We note that the waveform structure is a flexible hyperparameter that can be adapted. For instance, one may adjust the wave amplitude or frequency to moderate the intensity variation for shorter sequences. By scaling relative to the sequence length, we can ensure that each region of the generated sequence receives a proportional portion of the reference signals. This spatial modulation balances coverage and strength, allowing reference semantics to influence all regions of the generation while avoiding potential distortions caused by uniformly distributing the limited reference signals across a longer sequence. \\
\textbf{The Modified Update Rule:}
Finally, using \Cref{eq:down_up_sample} and \Cref{eq:my_wave}, we modify the update rule in line 8 of \Cref{alg:ilrr} as follows:
\begin{equation}
    h_{x_t}^{(k)} \leftarrow h_{x_t}^{(k)} + \mathbf{w} \cdot \hat{\Delta}^{(k)}
\end{equation}
This ensures that the reference signal is both spatially aligned via adaptive downsampling and dynamically distributed via the modulation vector $\mathbf{w}$.

\subsection{Controllable Steering Dynamics}
Our method introduces two key mechanisms that allow for tunable control over the steering behavior.
\textbf{(i) Steering Scale $\alpha$.} The scalar $\alpha$ determines the magnitude of semantic routing (\Cref{eq:ilrr_rule}), acting as a coefficient for the strength of the guidance. While a single global parameter is sufficient for general steering, our framework supports fine-grained control by allowing for layer-specific scaling. We can define a configuration vector $\boldsymbol{\alpha} = (\alpha_1, \dots, \alpha_{|L_s|})$ where each $\alpha_i$ corresponds to a specific injection layer $l_k \in L_s$. 
\textbf{(ii) Timesteps $T_s$.} We can further refine the steering by restricting the ILRR intervention to a specific subset of timesteps $T_s \subseteq \{1, \dots, T\}$, allowing us to apply semantic steering only during specific stages of the generation process (e.g., early semantic formation), leaving the remaining steps unconstrained.

\section{Experiments}
\subsection{Main Experiments}
\label{sec:main_experiments}
To evaluate ILRR for controllable generation, we examine its ability to steer DLMs toward semantic attributes, using references that exhibit the target attribute. We compare our method against state-of-the-art inference time particle-sampling-based baselines, including PG-DLM, FK-Steering and best-of-n \citep{dang2026inferencetime, singhal2025generalframeworkinferencetimescaling}. 

\begin{table}[t!]
\centering
\setlength{\tabcolsep}{5pt}
\label{tab:combined_results_final}
\resizebox{\columnwidth}{!}{%
\begin{tabular}{llcc}
\toprule
Model & Method & Toxicity ($\uparrow$) & Sentiment ($\uparrow$) \\
\midrule
\multirow{6}{*}{MDLM} 
 & best-of-$n$ & $1.9 \pm 0.4$ & $36.7 \pm 3.7$ \\
 & FK ($\phi=4$) & $0.8 \pm 0.2$ & $10.0 \pm 1.3$ \\
 & FK ($\phi=1$) & $3.8 \pm 1.0$ & $37.4 \pm 1.2$ \\
 & PG-DLM & $1.4 \pm 0.7$ & $23.8 \pm 2.2$ \\
 \cmidrule{2-4}
 & ILRR ($\alpha = 0.7$) & $ 4.7\pm 0.6$  & $43.4 \pm 1.3$ \\
 & ILRR ($\alpha = 0.8$) & $9.4 \pm 1.2$ & $51.3 \pm 4.1$ \\
 & ILRR ($\alpha = 1.0$) & $\textbf{15.9} \pm 2.0$ & $\textbf{65.2} \pm 2.5$ \\
\midrule
\midrule
\multirow{5}{*}{LLaDA} 
 & best-of-$n$ & $2.4 \pm 0.2$ & $48.2 \pm 2.9$ \\
 & FK & $9.0 \pm 1.5$ & $69.4 \pm 1.2$ \\
 & PG-DLM & $8.3 \pm 1.8$ & $66.6 \pm 1.0$ \\
 \cmidrule{2-4}
 & ILRR ($\alpha_1, \alpha_2 = 0.7, 0.75$) & $45.4 \pm 2.5$ & $97.7 \pm 0.3$ \\
 & ILRR ($\alpha_1, \alpha_2=0.8, 0.85$) & $60.7 \pm 0.7$ & $98.8 \pm 0.7$ \\
 & ILRR ($\alpha_1, \alpha_2 = 0.95, 1.0$) & $\textbf{71.2} \pm 1.2$ & $\textbf{100.0} \pm 0.0$ \\
\bottomrule
\end{tabular}%
}
\caption{Short generation results (length 50). Toxicity and positive sentiment accuracy (mean $\pm$ std) on MDLM and LLaDA across 3 random seeds.} 
\label{tab:combined_results_final}
\end{table}

\subsubsection{Experimental Setup}
\textbf{Datasets.} Following prior work \citep{dang2026inferencetime, singhal2025generalframeworkinferencetimescaling, han-etal-2023-ssd}, for each task we evaluate on the standard suite of 15 prefix prompts, generating 20 continuations per prompt. To ensure robust steering assessment, for each assessed attribute we utilize 5 distinct reference sequences per prompt, running 4 generations per reference-prompt pair. We repeat this setup across 3 random seeds. \\
\textbf{Metrics.} We evaluate steering effectiveness on toxicity and sentiment control. Following \citet{dang2026inferencetime}, we employ the following external classifiers:
\textbf{(i) Toxicity -} a RoBERTa-based toxicity classifier \citep{logacheva-etal-2022-paradetox} trained to detect harmful content (e.g., hate speech, insults). A generation is considered successful if it is classified as \textit{toxic}. While usually undesirable, toxicity is typically rare in model generations, making this task a rigorous benchmark for steering toward low-probability concepts.
\textbf{(ii) Sentiment -} a RoBERTa-based sentiment classifier \citep{barbieri-etal-2020-tweeteval} fine-tuned on the TweetEval benchmark. A generation is considered successful if it is classified as \textit{positive}.
For both tasks we report accuracy, defined as the percentage of generated sequences successfully classified as possessing the target attribute.

\subsubsection{Implementation Details}
\label{sec:implementation_details}
We apply ILRR to base DLMs, LLaDA-8B-Base \citep{nie2025largelanguagediffusionmodels} and MDLM \citep{sahoo2024simple}, integrating our method into their native generation loops.  We conduct two sets of experiments: (i) 50-length generation on LLaDA and MDLM, with length-matched references, used to evaluate our standard mechanism (\Cref{sec:ilrr}), and (ii) 512-length generation on MDLM, with approximately 60-token references, used to evaluate spatially modulated steering (\Cref{sec:longer_seq}). 
In all experiments, we operate under a compute budget measured by the number of function evaluations (\textit{NFE}), as defined in \citet{dang2026inferencetime}, under which our method corresponds to $\text{\textit{NFE}}=2 \cdot T$. For convenience, we normalize the \textit{NFE} value by the number of denoising steps $T$. We compare our approach against the baseline performance reported in \citet{dang2026inferencetime} at the comparable compute budget of normalized $\text{\textit{NFE}}=4$.
The full implementation and configuration details including our intervention layers, denoising and steering steps, scales, modulation parameters, datasets and reference curation are provided in \Cref{app:implementation_details}.

\subsubsection{Main Results} \Cref{tab:combined_results_final} reports steering accuracy for 50-token generation on LLaDA and MDLM, showing ILRR consistently outperforms all inference-time baselines in the constrained compute budget. On LLaDA, ILRR reaches $71.2\%$ toxicity and $100.0\%$ sentiment accuracy, compared to $9.0\%$ and $69.4\%$ for the strongest baselines. On MDLM, ILRR improves toxicity from $3.8\%$ to $15.9\%$ and sentiment from $37.4\%$ to $65.2\%$. \Cref{tab:long_sequence_results} details the results for 512-token generation on MDLM using spatially modulated steering, showing that ILRR effectively improves toxicity accuracy from $3.0\%$ to $12.1\%$ and sentiment accuracy from $26.0\%$ to $58.4\%$, an improvement of $32\%$ points.

\begin{table}[t!]
\centering
\small
\begin{tabular*}{\columnwidth}{@{\extracolsep{\fill}}lcc}
\toprule
Method & Toxicity ($\uparrow$) & Sentiment ($\uparrow$) \\
\midrule
best-of-$n$ & 1.0 & 23.0 \\
FK ($\phi=4$) & 0.0 & 7.3 \\
FK ($\phi=1$) & 3.0 & 26.0 \\
PG-DLM & 1.7 & 17.3 \\
\midrule
ILRR ($\alpha = 0.7$) & $3.8 \pm 0.5$ & $43.7 \pm 2.5$ \\
ILRR ($\alpha = 0.8$) & 5.7 $\pm$ 2.0 & 48.2 $\pm$ 1.9 \\
ILRR ($\alpha = 1.0$) & \textbf{12.1} $\pm$ 2.0 & \textbf{58.4} $\pm$ 3.4 \\
\bottomrule
\end{tabular*}
\caption{Long generation results (length 512). Toxicity and positive sentiment accuracy on MDLM using spatially modulated steering across 3 random seeds.} 
\label{tab:long_sequence_results}
\end{table}

\subsection{Generalization and Broad Applicability}
\label{sec:generalization}
To evaluate ILRR's robustness, we conduct additional experiments on LLaDA, testing (i) whether the same hyperparameter configuration (\Cref{app:implementation_details}) transfers to additional control objectives of detoxification and topic control, and (ii) whether small pools of reused references can steer generations of diverse prompts, reflecting a practical setting in which references are curated once for a desired attribute or domain and reused across different conditioned generations.
For detoxification, we use the 791 most toxic prompts from RealToxicityPrompts \citep{gehman2020realtoxicityprompts} and steer toward non-toxic continuations using only 20 global safe references. Each prompt is paired with its top-3 cosine-similar references, and we generate 4 40-length continuations per pair with toxicity evaluated using \citet{logacheva-etal-2022-paradetox}. For topic control, we follow \citet{dathathri2020plugplaylanguagemodels} and steer their standard suite of 20 topic control prompts toward Space, Religion, and Science using 5 global references per topic, generating 4 50-length continuations per prompt-reference pair. Since dedicated classifiers are not naturally available for such open-ended topics, we evaluate topic relevance using an LLM-as-a-judge (GPT-4o-mini), prompted to assign a 1-5 relevance score.
As shown in \Cref{tab:generalization}, ILRR substantially reduces toxicity and achieves high topic relevance, suggesting that the method can support effective zero-shot steering - small pools of references can be reused to guide diverse prompts, without task-specific tuning.\\
\begin{table}[t]
    \centering
    \small
    \begin{subtable}[t]{\columnwidth}
        \centering
        \begin{tabular}[t]{lc}
        \toprule
        Method & Toxicity (\%) ($\downarrow$) \\
        \midrule
        Base Model & 40.9 \\
        best-of-$n$ ($n=2$) & 24.9 \\
        best-of-$n$ ($n=4$) & 13.2 \\
        \midrule
        ILRR ($\alpha_1, \alpha_2 = 0.7,0.75$) & 10.7 \\
        ILRR ($\alpha_1, \alpha_2 = 0.8,0.85$) & \textbf{8.0} \\
        \bottomrule
        \end{tabular}
        \caption{Detoxification results on the most toxic RealToxicityPrompts examples.}
        \label{tab:detoxification}
    \end{subtable}
    
    \vspace{0.4cm} 
    \begin{subtable}[t]{\columnwidth}
        \centering
        \begin{tabular}[t]{lcc}
        \toprule
        \multicolumn{3}{c}{Relevance $\ge 4$ (\%) ($\uparrow$)} \\
        \midrule
        Topic / Scale & $(0.7,0.75)$ & $(0.8,0.85)$ \\
        \midrule
        Space & 94.3 & \textbf{97.5} \\
        Religion & 88.8 & \textbf{94.0} \\
        Science & 84.3 & \textbf{89.0} \\
        \bottomrule
        \end{tabular}
        \caption{ILRR topic control results. Topic relevance is evaluated using GPT-4o-mini as-a-judge.}
        \label{tab:topic_control}
    \end{subtable}
    
    \caption{Generalization averaged results (LLaDA) using small pools of globally reused references.}
    \label{tab:generalization}
\end{table}
\\Overall, the results indicate that iterative latent intervention is an effective mechanism for steering DLMs. ILRR transfers reference-derived guidance across all tasks and length settings, and enables efficient steering under a relatively low computational budget compared to the sampling-based baselines. We attribute this effectiveness to the rich sequence-wide semantic information encoded in DLM hidden states; by gradually aligning these representations with reference-derived signals, the generation trajectory is biased toward the desired semantic direction. \\We additionally provide the score distributions given by the external classifiers in \Cref{app:score_distributions}, and qualitative examples in \Cref{tab:qualitative_examples}.

\section{Analysis and Ablation}
\label{sec:analysis_and_ablation}
To better understand the effect of each component of our framework, we analyze how key parameters impact the steering effectiveness of ILRR while keeping all other parameters fixed. Further analysis details are provided in \Cref{app:analysis_and_ablation}.

\begin{figure}[h]
    \centering
    \begin{subfigure}[h]{0.49\columnwidth}
        \centering
        \includegraphics[trim=0.4cm 0.0cm 0.0cm 0.0cm, clip, width=\linewidth]{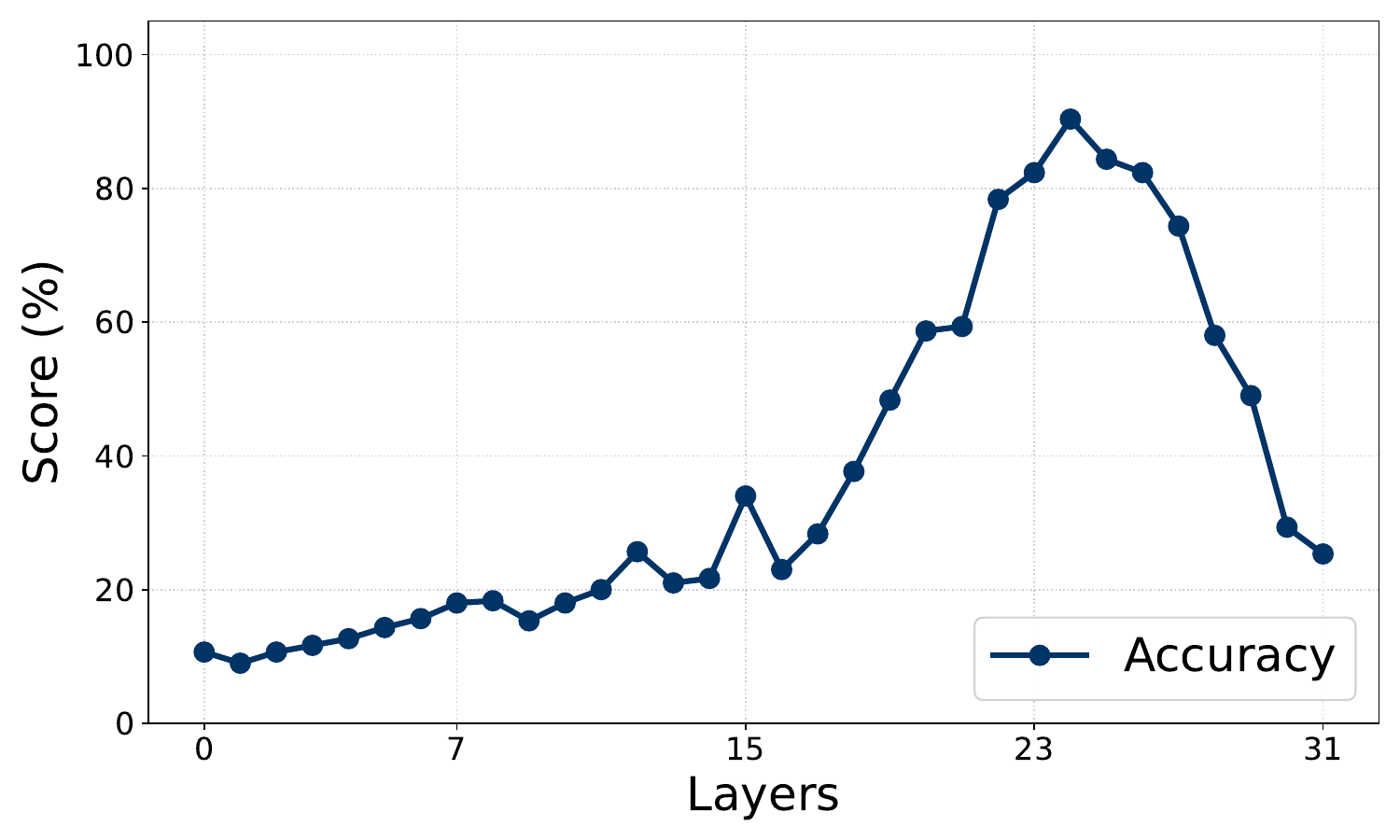}
        \caption{}
        \label{fig:layer_single}
    \end{subfigure}
    \hfill
    \begin{subfigure}[h]{0.49\columnwidth}
        \centering
        \includegraphics[trim=0.4cm 0.4cm 0.0cm 0.0cm, clip, width=\linewidth]{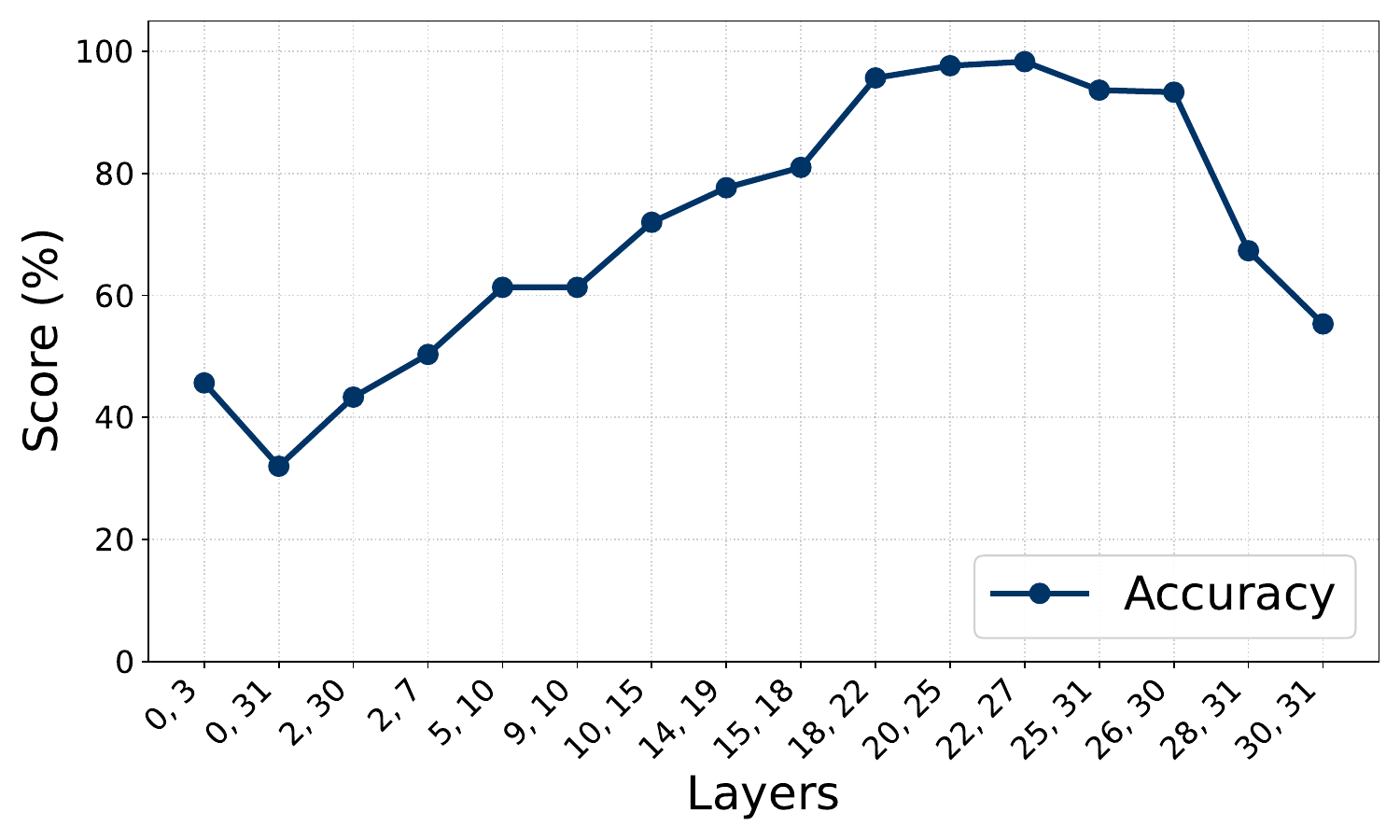}
        \caption{}
        \label{fig:layer_pairs}
    \end{subfigure}
    
    \vspace{0.2cm} 
    \begin{subfigure}[h]{0.48\columnwidth}
        \centering
        \includegraphics[trim=0.2cm 0.2cm 0.2cm 0.2cm, clip, width=\linewidth]{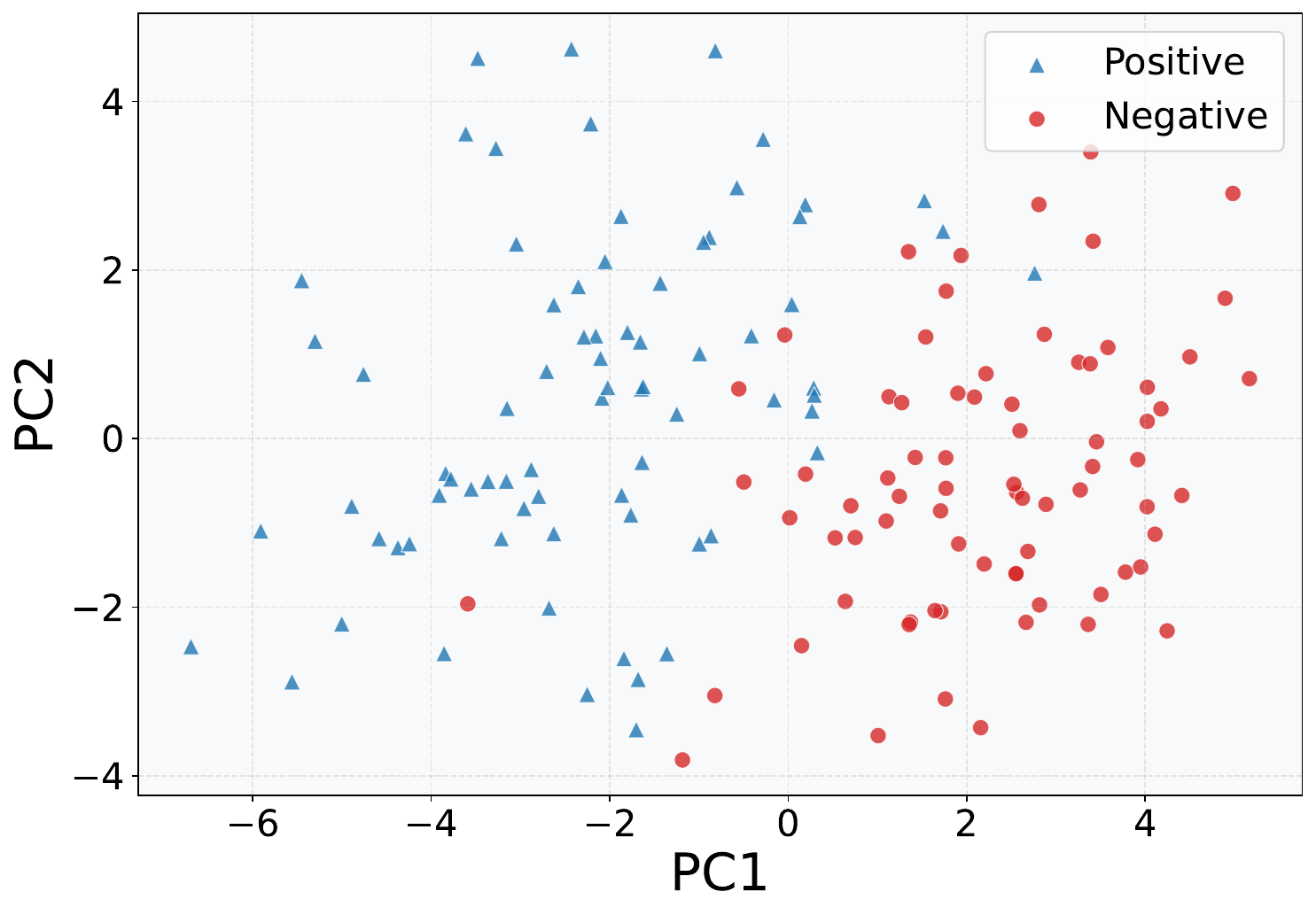}
        \caption{}
        \label{fig:pca_layer_6}
    \end{subfigure}
    \hfill
    \begin{subfigure}[h]{0.48\columnwidth}
        \centering
        \includegraphics[trim=0.2cm 0.2cm 0.2cm 0.2cm, clip, width=\linewidth]{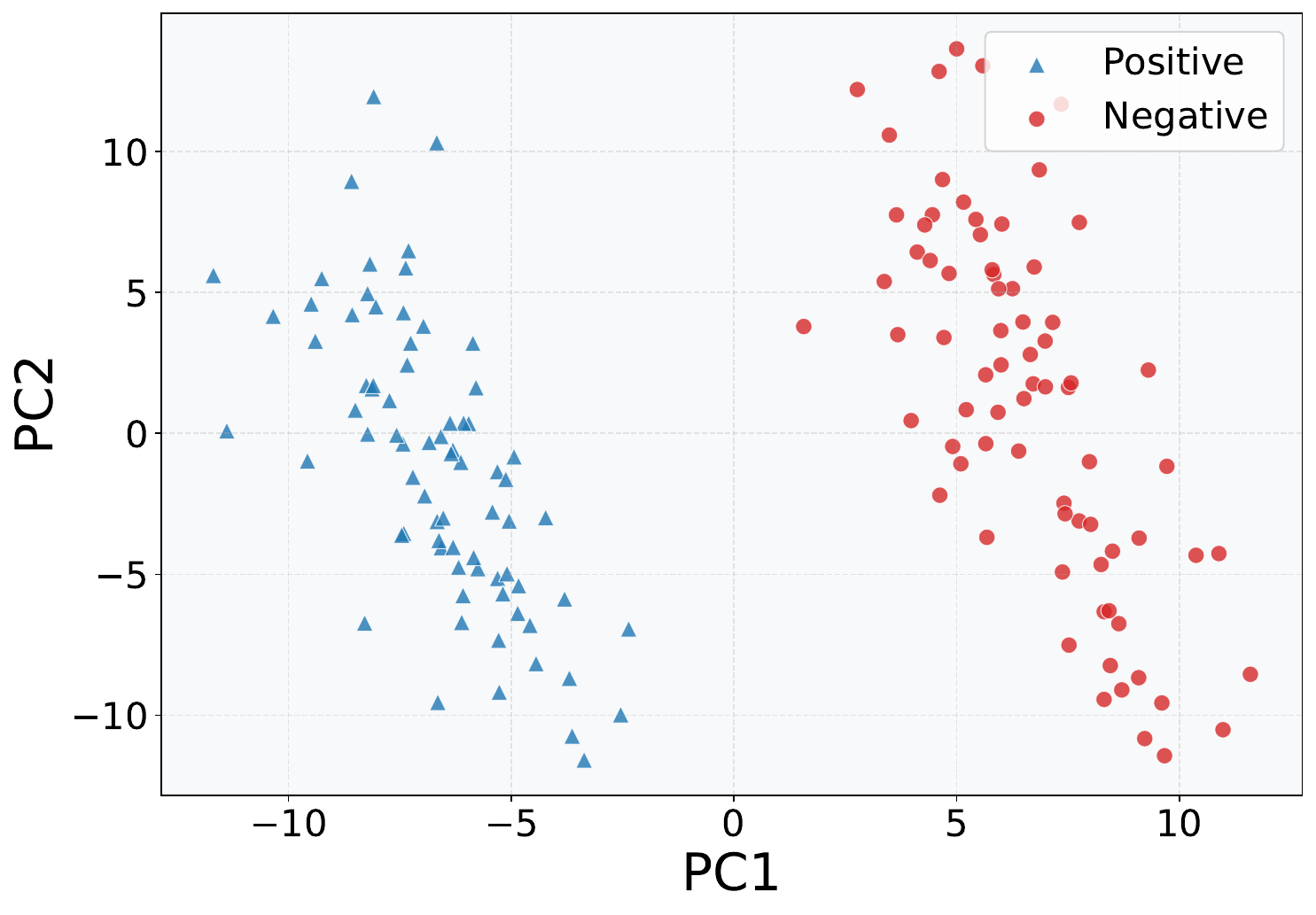}
        \caption{}
        \label{fig:pca_layer_20}
    \end{subfigure}
    
    \vspace{0.2cm} 
    \begin{subfigure}[h]{0.65\columnwidth}
        \centering
        \includegraphics[trim=0.2cm 0.2cm 0.2cm 0.2cm, clip, width=\linewidth]{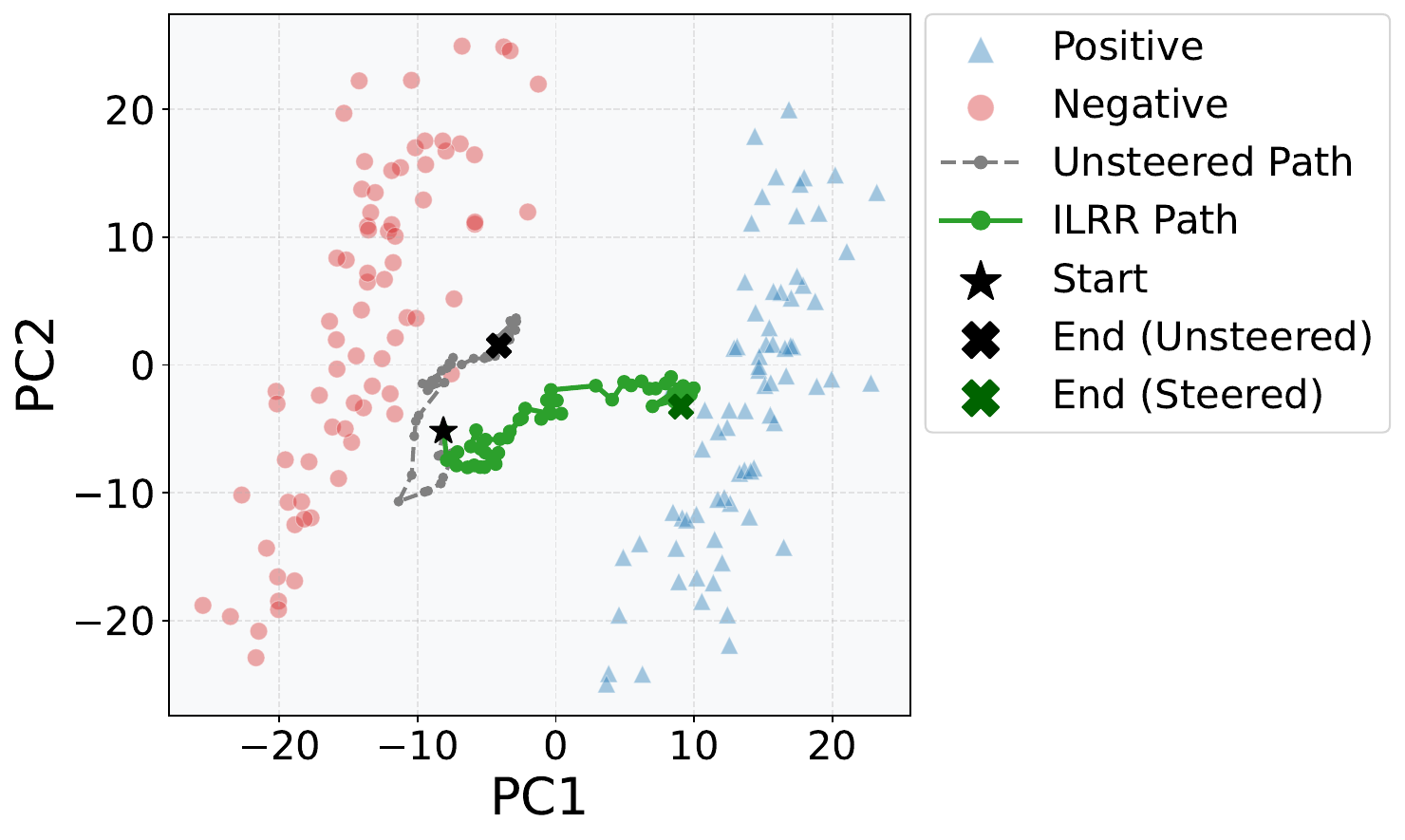}
        \caption{}
        \label{fig:trajectory_layer_25}
    \end{subfigure}
    
    \caption{Layers and latent geometry. (a) Accuracy across single model layers. (b) Accuracy across pairs of model layers. (c, d) PCA projections of latent representations for positive and negative sequences at layers 6 and 20, respectively. (e) Tracking the steered vs. unsteered generation trajectories via PCA at layer 25.}
    \label{fig:layer_geometry_analysis}
\end{figure}
\paragraph{Layer selection and latent geometry} (\Cref{app:layers_ablation}).
\Cref{fig:layer_single,fig:layer_pairs} show that steering is most effective in mid-to-late layers, while early-layer interventions are consistently weaker. \Cref{fig:pca_layer_6,fig:pca_layer_20,fig:trajectory_layer_25} provide a geometric explanation for this trend: representations are more entangled in early layers but become separable in mid-to-late layers, where ILRR can redirect the denoising trajectory toward the target semantic region.

\begin{figure}[h]
    \centering
    \begin{subfigure}[t]{0.5\columnwidth}
        \centering
        \includegraphics[width=\linewidth]{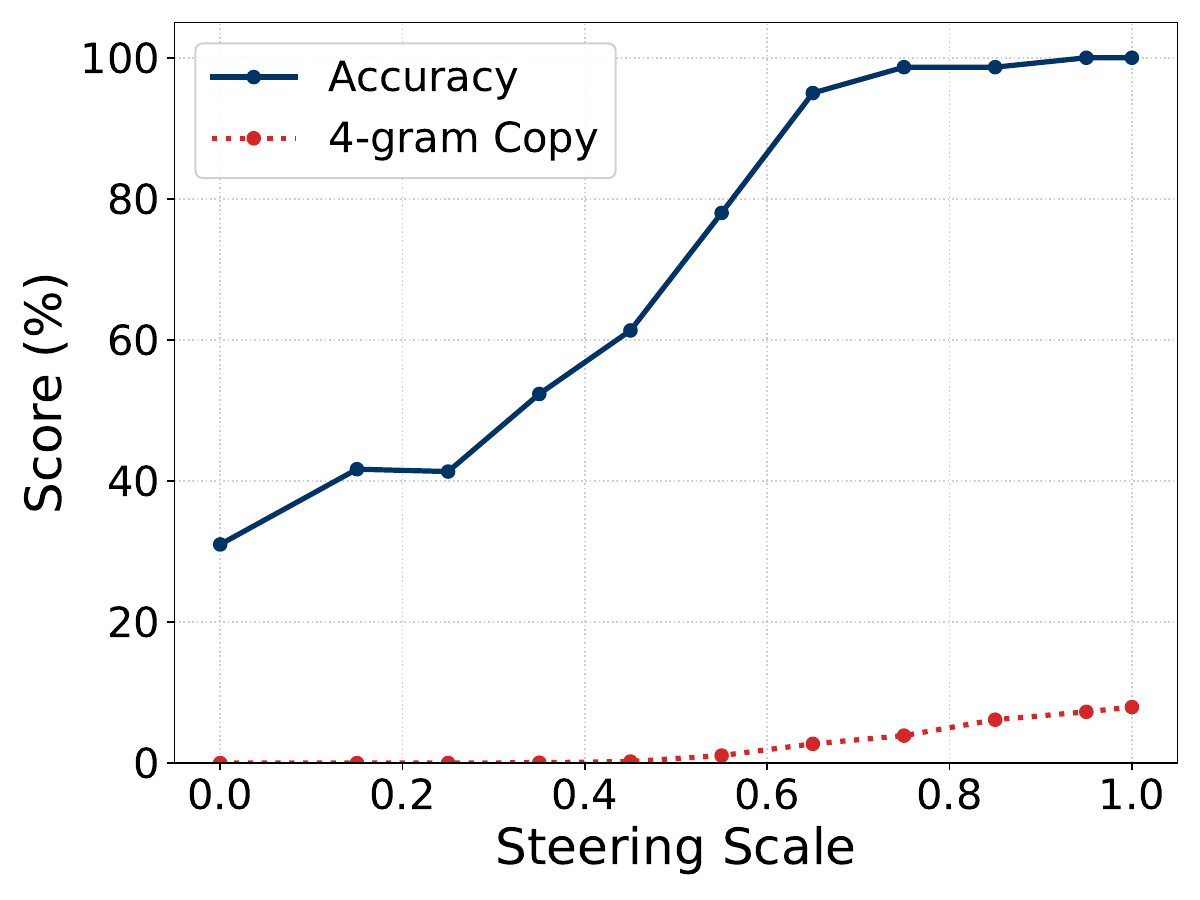}
        \caption{}
        \label{fig:steering_scale}
    \end{subfigure}
    \hfill
    \begin{subfigure}[t]{0.41\columnwidth}
        \centering
        \includegraphics[width=\linewidth]{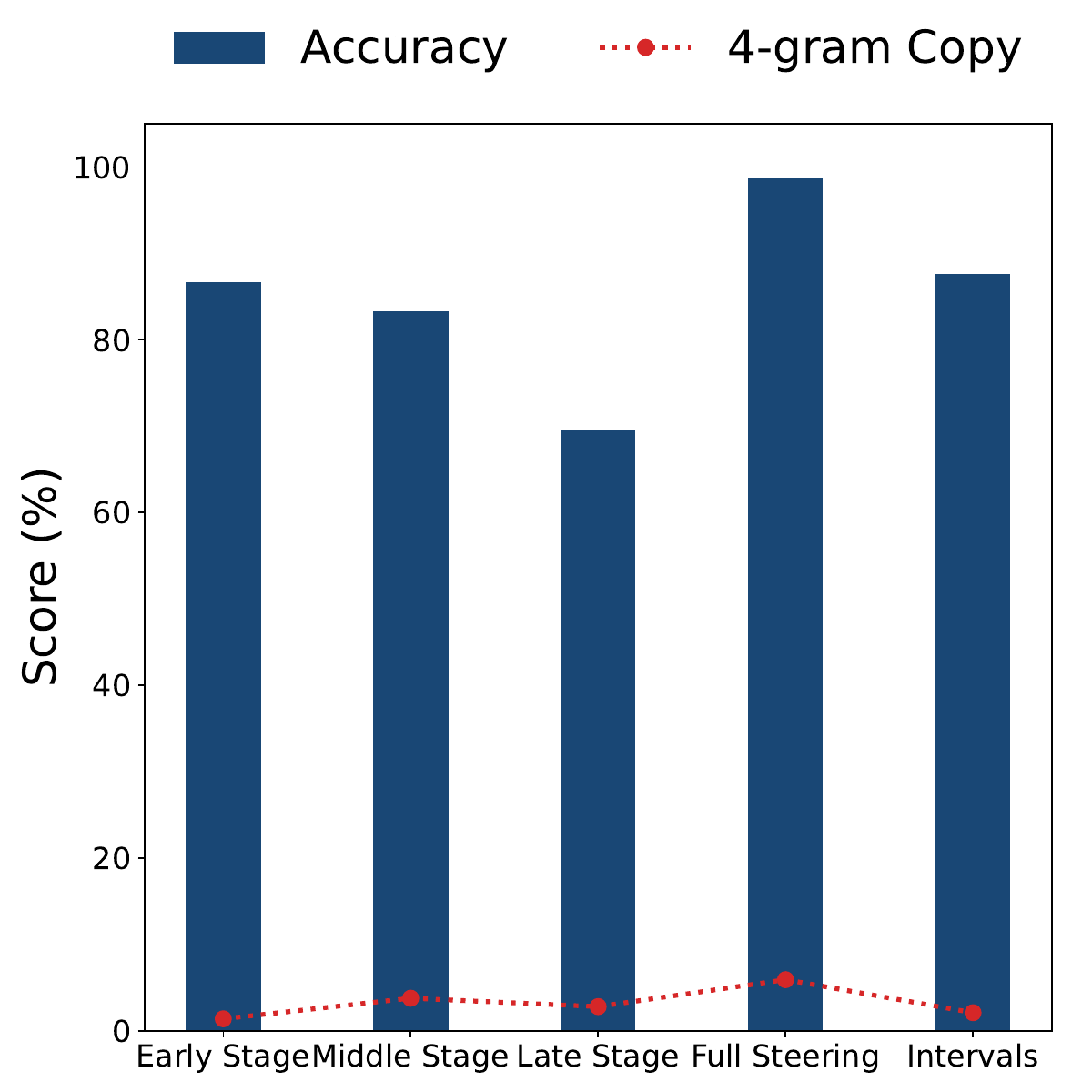}
        \caption{}
        \label{fig:steering_steps}
    \end{subfigure}
    \caption{Tunability analysis. (a) Accuracy compared to 4-gram overlap with the reference text across steering scales $\alpha$. (b) Accuracy compared to 4-gram overlap with the reference text across steering step sets $T_s$.}
    \label{fig:tunability_analysis}
\end{figure}

\begin{table}[h]
\centering
\setlength{\tabcolsep}{3pt}
\resizebox{\columnwidth}{!}{%
\begin{tabular}{lcccc}
\toprule
Reference type & Acc (\%)($\uparrow$) & Flu. ($\uparrow$) & Dist-1/2/3 ($\uparrow$) & PPL ($\downarrow$) \\
\midrule
Original & \textbf{97.3} & \textbf{4.2} & .42 / .74 / .84 & 18.3 \\
\midrule
Sparse (25\% word drop) & 86.7 & 3.0 & .31 / .56 / .66 & 12.5 \\
Noise (50\% random words) & 69.3 & 2.6 & .33 / .60 / .73 & 13.3 \\
Sparse (50\% word drop) & 45.3 & 1.8 & .14 / .26 / .32 & 5.4 \\
\midrule
Chunk-shuffled (10-words) & 96.0 & 3.7 & .40 / .72 / .85 & 17.0 \\
Chunk-shuffled (5-words) & 86.7 & 3.5 & .37 / .66 / .79 & 14.9 \\
Chunk-shuffled (2-words) & 73.3 & 2.8 & .32 / .61 / .75 & 13.8 \\
Fully Shuffled & 61.3 & 2.2 & .29 / .57 / .73 & 14.4 \\
\midrule
Highly repetitive & 84.0 & 2.1 & .13 / .23 / .29 & \textbf{3.5} \\
Midway sentiment shift & 46.7 & 3.9 & \textbf{.44 / .76 / .86} & 20.7 \\
Grammatically correct gibberish & 37.3 & 3.4 & .36 / .57 / .68 & 6.8 \\
\bottomrule
\end{tabular}%
}
\caption{The impact of reference corruption, structural shuffling and semantic degradation on the generation. Flu. denotes GPT-4o fluency score on a 1–5 scale.}
\label{tab:ref_quality}
\end{table}

\paragraph{Tunability} (\Cref{app:tunability}).
\Cref{fig:steering_scale} shows that steering intensity increases consistently with $\alpha$, while 4-gram overlap with the reference remains low even at the highest scale, indicating that stronger control arises from semantic alignment rather than heavy verbatim reuse of the reference. \Cref{fig:steering_steps} shows that ILRR is effective in various denoising stages, with early intervention outperforming late, indicating that coarse semantic control is more influential early in generation.

\paragraph{Reference quality} (\Cref{app:reference_quality_ablation}).
\Cref{tab:ref_quality} shows that effective references require coherent, consistent semantics and sufficient lexical diversity, while corrupted, shuffled, inconsistent, or repetitive references degrade control and quality.

\begin{figure}[h]
    \centering
    \includegraphics[width=0.52\columnwidth]{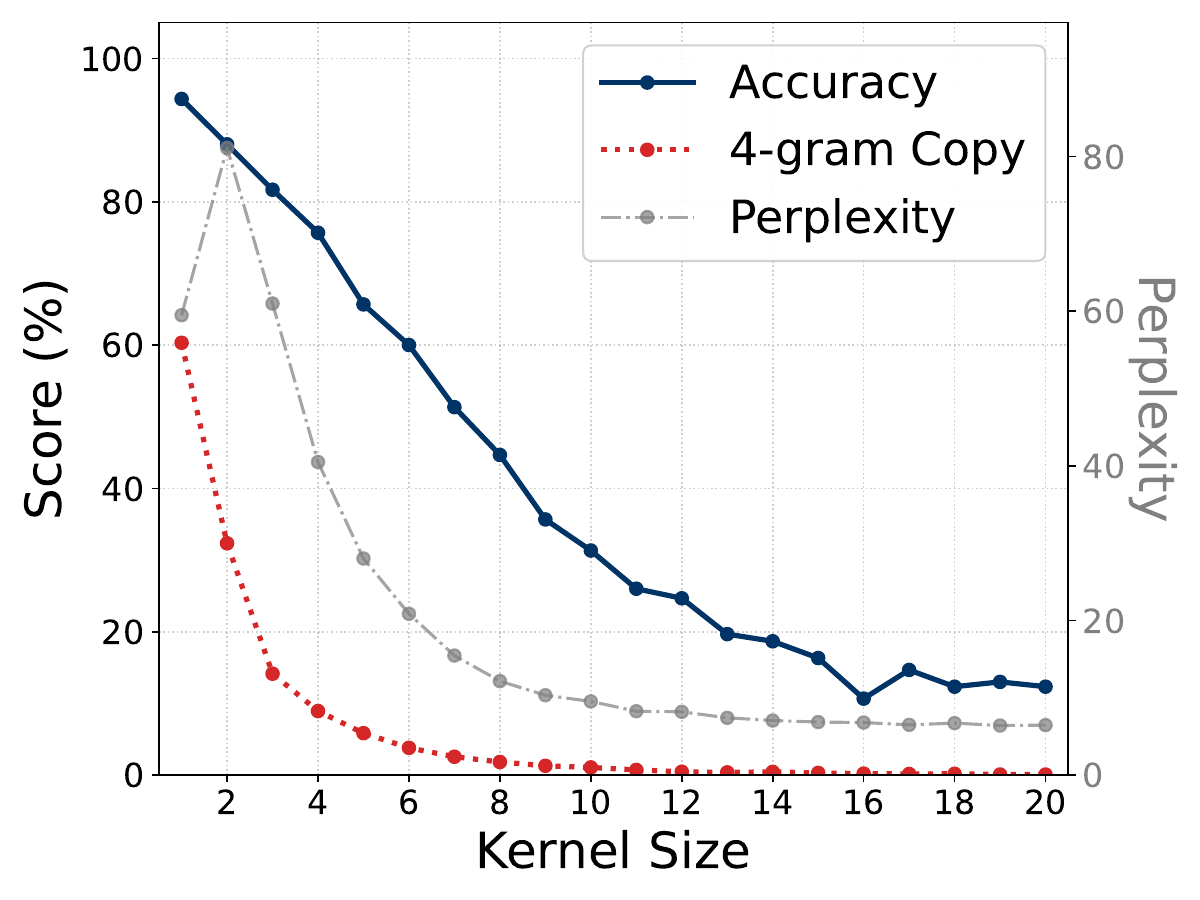}
    \caption{Accuracy across pooling kernel sizes $k$ compared to PPL and 4-gram overlap with the reference.}
    \label{fig:kernels}
\end{figure}
\paragraph{Kernel size for semantic extraction} (\Cref{app:kernel_ablation}).
\Cref{fig:kernels} shows that small kernels overemphasize local token-level alignment, increasing both 4-gram overlap and PPL. Moderate kernels maintain strong steering while reducing these effects, supporting the role of pooling as a coarse semantic extractor that suppresses fine-grained reference details.

\begin{figure}[h]
    \centering
    
    \begin{subfigure}[b]{\columnwidth}
        \centering
        \includegraphics[width=0.65\linewidth]{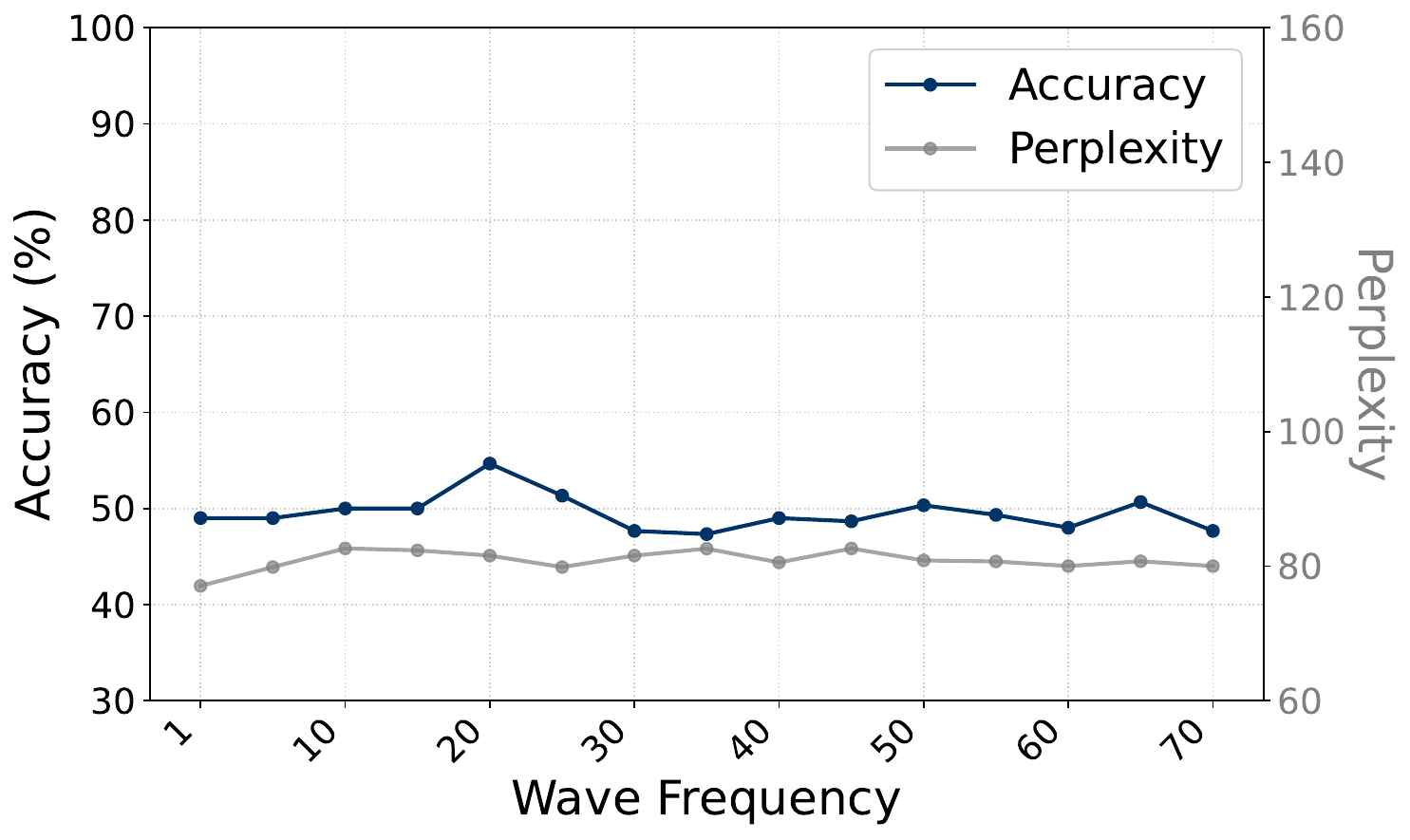}
        \caption{}
        \label{fig:modulated_steering}
    \end{subfigure}
    
    \vspace{0.4cm}
    
    \begin{subfigure}[b]{\columnwidth}
        \centering
        \small 
        \begin{tabular}{lcc}
        \toprule
        Waveform Shape & Acc (\%) ($\uparrow$) & PPL ($\downarrow$) \\
        \midrule
        Cosine (Ours) & 50.0 & 82.3 \\
        Linear Ramp & 51.6 & 80.2 \\
        Square & 50.6 & 91.0 \\
        Concave Parabola & 61.3 & 100.8 \\
        \midrule
        Sigmoid & 39.0 & 85.9 \\
        Gaussian & 33.7 & 77.4 \\
        Exponential Decay & 22.3 & \textbf{49.2} \\
        Exponential Decay - Wave & 21.3 & 52.6 \\ 
        Exponential Wave & 18.7 & 52.4 \\
        Exponential & 18.0 & 53.0 \\
        \midrule
        Uniform (Constant $\alpha$) & \textbf{72.0} & 142.4 \\
        \bottomrule
        \end{tabular}
        \caption{}
        \label{tab:waveforms}
    \end{subfigure}
    
    \caption{Spatial Modulation analysis. (a) Accuracy and PPL across varying wave frequencies $f$. (b) Impact of waveform shape.}
    \label{fig:spatial_modulation_combined}
\end{figure}

\paragraph{Spatial Modulation}
(\Cref{app:spatially_modulated_steering_ablation}). \Cref{tab:waveforms} shows that broad alternating waveforms are most effective for balancing controllability and likelihood, whereas uniform steering is overly aggressive and substantially worsens PPL. \Cref{fig:modulated_steering} further shows stable performance across a wide range of frequencies, indicating that effectiveness comes from spatially distributing the steering signal rather than from careful tuning of $f$.

\section{Conclusion}
We introduced Iterative Latent Representation Refinement (ILRR), a learning-free framework for steering DLMs by injecting reference-derived semantic guidance into the evolving denoising trajectory. We showed that ILRR achieves effective plug-and-play control with low computational overhead, outperforming baselines under comparable budgets, extending to long-form generation and remaining effective in broader control settings. Overall, our results highlight ILRR as a practical low-compute approach for controlled DLM generation, and position reference-based latent intervention as an efficient paradigm for steering non-AR generation.

\section{Limitations}
Although ILRR demonstrates effective inference-time steering across multiple architectures and control tasks, our framework considers exclusively text-based references; whether other reference modalities can similarly guide DLM generation remains outside the scope of the present study. Additionally, while extending ILRR to simultaneously compose multiple reference signals for multi-attribute control is a natural direction, the precise mechanisms for balancing and composing multiple-source signals require further investigation, which we leave for future work. Finally, this work focuses on establishing and validating ILRR through coarse semantic attribute control tasks, while extending it to complex downstream tasks such as reasoning entails further investigation into their latent interactions and remains a key direction for future research.

\section{Ethics Statement}
This work explores novel and efficient approaches for controlling diffusion language models at inference time, with the aim of providing deeper insight into model capabilities and promoting further research in controllable text diffusion generation. At the same time, we recognize the potential risk of misuse or malicious exploitation inherent to such controllable generation techniques. We encourage the appropriate use of this work and believe that future applications could utilize our findings to focus on advancing safer and more responsible generative technologies.
\section{AI Use Disclosure}
AI tools were used to edit and polish the writing of this manuscript.

\bibliography{custom}

\appendix
\label{sec:appendix}

\newpage
\section{Implementation Details}
\label{app:implementation_details}

\subsection{General Implementation Details}
We apply ILRR to two discrete diffusion base models, LLaDA-8B-Base \citep{nie2025largelanguagediffusionmodels} and MDLM \citep{sahoo2024simple}, integrating our method into their native generation loops. In all experiments, we operate under a compute budget measured by the number of function evaluations (\textit{NFE}), as defined in \citet{dang2026inferencetime}, under which our method corresponds to $\text{\textit{NFE}}=2 \cdot T$. For convenience, we normalize the \textit{NFE} value by the number of denoising steps $T$. 
We conduct all experiments on a single NVIDIA RTX-4090 GPU. 
Our code is available at \url{https://github.com/edenavr555/ILRR-Steering}.

\subsection{Setup and Hyperparameters}
\paragraph{Shorter sequence generation.} To evaluate our standard ILRR mechanism (\Cref{sec:ilrr}), we generate sequences of length 50 using text references of equal length for both base models. For LLaDA, we follow the model’s standard inference configuration, where the number of denoising steps typically does not exceed the generation length. Accordingly, all methods run with $T=50$ denoising steps. We apply steering at layers $L_s \in \{20, 25\}$ throughout the entire denoising process ($T_s \in \{1, \dots, 50\}$). We report results at steering scales $(\alpha_1, \alpha_2) \in \{(0.7, 0.75), (0.8, 0.85), (0.95, 1.0)\}$. In this setting, our method operates at a normalized $\textit{NFE}=2$. For MDLM, baseline methods employ $T=1024$ denoising steps, whereas we run our method with $T=2048$ steps in order to reach the compute budget of normalized $\textit{NFE}=4$. Steering is applied at layer $L_s=\{9\}$ during the first half of the denoising process ($T_s \in \{1, \dots, 900\}$), and we report results at steering scales $\alpha \in \{0.7, 0.8, 1.0\}$.

\paragraph{Long sequence generation.} To evaluate our spatially modulated steering mechanism (\Cref{sec:longer_seq}), we generate longer sequences of length 512 with approximately 60-token-long references, using the MDLM base model. Baseline methods operate with $512$ denoising steps, while we employ $T=1024$ steps to reach the comparable compute budget of $\textit{NFE}=4$. Steering is applied at layer $L_s=\{9\}$ for $T_s \in \{1, \dots, 900\}$, using a cosine modulation waveform with frequency $f=7$. We report results at steering scales $\alpha \in \{0.7, 0.8, 1.0\}$. For compute comparison in this setting, we normalize \textit{NFE} by the baseline step count of $512$. Under this normalization, our method operates at a compute budget of normalized $\textit{NFE}=4$.

\paragraph{}In all experiments, we use a pooling kernel size of 6. 

\subsection{Datasets and Classifiers}
For the main experiments (\Cref{sec:main_experiments}), we follow prior work \citep{dang2026inferencetime, singhal2025generalframeworkinferencetimescaling, han-etal-2023-ssd} and steer the standard suite of 15 prefix prompts (\Cref{fig:prompts_main}). For our generalization experiments (\Cref{sec:generalization}), we follow \citet{dathathri2020plugplaylanguagemodels} and steer their suite of 20 prefix prompts (\Cref{fig:prompts_topic}) for topic control, and use the 791 most toxic prompts from the RealToxicityPrompts dataset (\citet{gehman2020realtoxicityprompts}, \url{https://huggingface.co/datasets/allenai/real-toxicity-prompts}) for detoxification. 
The external classifiers used to evaluate sentiment and toxicity throughout all experiments are available at \url{https://huggingface.co/cardiffnlp/twitter-roberta-base-sentiment} and \url{https://huggingface.co/s-nlp/roberta_toxicity_classifier}.

\begin{figure}[h]
    \centering
    \begin{subfigure}{\linewidth}
        \begin{tcolorbox}[colback=gray!5!white,colframe=gray!75!black,boxrule=0.5pt,arc=4pt,width=\linewidth,nobeforeafter]
            \textit{"Once upon a time", "The book", "The chicken", "The city", "The country", "The horse", "The lake", "The last time", "The movie", "The painting", "The pizza", "The potato", "The president of the country", "The road", "The year is 1910"}
        \end{tcolorbox}
        \caption{Prefix prompts used for the main experiments.}
        \label{fig:prompts_main}
    \end{subfigure}
    
    \vspace{0.4cm}
    \begin{subfigure}{\linewidth}
        \begin{tcolorbox}[colback=gray!5!white,colframe=gray!75!black,boxrule=0.5pt,arc=4pt,width=\linewidth,nobeforeafter]
            \textit{"In summary", "This essay discusses", "Views on", "The connection", "Foundational to this is", "To review", "In brief", "An illustration of", "Furthermore", "The central theme", "To conclude", "The key aspect", "Prior to this", "Emphasised are", "To summarise", "The relationship", "More importantly", "It has been shown", "The issue focused on", "In this essay"}
        \end{tcolorbox}
        \caption{Prefix prompts used for the topic control experiments.}
        \label{fig:prompts_topic}
    \end{subfigure}
    
    \caption{Prefix prompts used across our control tasks.}
    \label{fig:prefix_prompts_datasets}
\end{figure}

\begin{figure*}[t!]
    \centering
    
    \begin{tcolorbox}[
        title=\textbf{Generation Prompt Template}, 
        colback=gray!5!white, 
        colframe=gray!75!black, 
        boxrule=0.5pt, 
        arc=4pt, 
        width=\textwidth, 
        nobeforeafter
    ]
        \textit{Generate a natural, coherent exemplar text of approximately \{$l^*$\} words that strongly and clearly exemplifies the following attribute: \{$a^*$\}. To ensure the text is effective, please follow these guidelines: 1. Natural and Coherent: The text should read naturally and fluently. You are free to choose the content, as long as the desired attribute is clearly demonstrated. 2. Global Consistency: Maintain the target attribute steadily throughout the entire text. Please avoid any sudden shifts, contrasts, or contradictions in tone and meaning. 3. Lexical Diversity: Express the core idea using a rich and varied vocabulary. Avoid echoing the exact same words or phrasing repetitively.}
    \end{tcolorbox}
    
    \vspace{0.4cm} 
    \begin{tcolorbox}[
        title=\textbf{Evaluation Prompt Template}, 
        colback=gray!5!white, 
        colframe=gray!75!black, 
        boxrule=0.5pt, 
        arc=4pt, 
        width=\textwidth, 
        nobeforeafter
    ]
        \textit{You are an impartial and rigorous judge evaluating a candidate text based on how well it embodies a specific semantic attribute. Target Attribute: \{$a^*$\}. Candidate Text: \{$y_{cand}$\}. Evaluate how well the candidate text exemplifies the target attribute on a scale of 1 to 5, strictly using the following rubric: 1 - Poor: The text does not exhibit the target attribute, contains incorrect gibberish, or features major contradictory shifts in meaning. 2 - Fair: The attribute is weakly present. The text suffers from poor fluency, extreme repetition, or noticeable shifts in tone. 3 - Good: The attribute is clearly present, but the text lacks lexical diversity or has minor inconsistencies in its semantic focus. 4 - Very Good: The text clearly and consistently demonstrates the attribute from start to finish, maintaining good fluency and varied vocabulary. 5 - Excellent: The text is a flawless, highly coherent, and lexically diverse representation of the attribute without any deviation in tone or quality. You may provide a short explanation for your score. You MUST format your response as a valid JSON object exactly like the following structure: \{"score": <integer from 1 to 5>, "explanation": <short string explaining the reasoning behind the score>\}}
    \end{tcolorbox}
    
    \caption{Prompt templates used in our reference curation pipeline.}
    \label{fig:all_prompts}
\end{figure*}

\subsection{Reference Curation}
\label{app:reference_curation}
For all experiments, reference pools were obtained using Gemini 3.1 Pro, prompted and validated to exhibit the target semantics while keeping the text natural and diverse.

We employ a formal reference curation pipeline, which operates purely as a one-time pre-processing step, separate from the generation pipeline. We utilize prompt templates for generation and evaluation (\Cref{fig:all_prompts}), populated with the desired attribute $a^*$ and target length $l^*$, and operate as follows:

\begin{enumerate}
    \item \textbf{Generation:} Populate the generation template with the target attribute $a^*$ and desired length $l^*$ to create the prompt $P_{gen}$. Prompt a generator model $M_{gen}$ with $P_{gen}$ to produce a candidate reference text $y_{cand}$.
    \item \textbf{Evaluation:} Populate the evaluation template with $a^*$ and the candidate text $y_{cand}$ to create the prompt $P_{eval}$. Pass $P_{eval}$ to an evaluator model $M_{eval}$ to score how well the candidate reflects the target attribute.
    \item \textbf{Validation and Iteration:} If the candidate's score meets a predefined threshold $\tau$, accept $y_{cand}$ as a valid reference $y_0$. If not, repeat the process until a valid reference is found.
\end{enumerate}

\section{Score Distributions}
\label{app:score_distributions}

To provide deeper insight into ILRR's steering effectiveness, we additionally provide the score distributions given by the external classifiers across our experiments, where higher scores indicate greater toxicity ($\uparrow$) and positive sentiment ($\uparrow$). \Cref{tab:score_distributions} displays the percentage of generations falling into distinct score ranges across varying steering scales, including the unsteered baseline (scale $0.0$), demonstrating the distributional shifts that grow with steering scale.

\begin{table*}[h]
\centering
\small
\begin{tabular}{ll c ccccc}
\toprule
\multirow{2}{*}{{Model}} & \multirow{2}{*}{{Task}} & \multirow{2}{*}{{Scale $\alpha$}} & \multicolumn{5}{c}{{Classifier Score Ranges (\%)}} \\
\cmidrule(lr){4-8}
& & & {[0.0--0.2]} & {(0.2--0.4]} & {(0.4--0.6]} & {(0.6--0.8]} & {(0.8--1.0]} \\
\midrule
\multirow{16}{*}{{MDLM}} 
& \multirow{4}{*}{Sentiment (\Cref{tab:combined_results_final})} 
  & $0.0$ & 86.0 & 7.2 & 2.4 & 2.4 & 1.9 \\
& & $0.7$ & 36.6 & 14.9 & 13.0 & 13.7 & 21.9 \\
& & $0.8$ & 27.1 & 14.3 & 13.8 & 18.0 & 26.8 \\
& & $1.0$ & 16.2 & 12.9 & 14.0 & 20.3 & 36.6 \\
\cmidrule{2-8}
& \multirow{4}{*}{Toxicity (\Cref{tab:combined_results_final})} 
  & $0.0$ & 99.1 & 0.1 & 0.2 & 0.1 & 0.4 \\
& & $0.7$ & 94.3 & 0.8 & 0.3 & 0.3 & 4.2 \\
& & $0.8$ & 89.9 & 0.7 & 0.2 & 0.3 & 8.9 \\
& & $1.0$ & 83.3 & 0.7 & 0.2 & 0.8 & 15.0 \\
\cmidrule{2-8}
& \multirow{4}{*}{Sentiment (\Cref{tab:long_sequence_results})} 
  & $0.0$ & 69.9 & 17.3 & 7.9 & 3.4 & 1.4 \\
& & $0.7$ & 14.7 & 30.6 & 28.8 & 20.1 & 5.9 \\
& & $0.8$ & 12.1 & 27.6 & 30.3 & 22.8 & 7.2 \\
& & $1.0$ & 5.8 & 22.7 & 34.7 & 28.3 & 8.6 \\
\cmidrule{2-8}
& \multirow{4}{*}{Toxicity (\Cref{tab:long_sequence_results})} 
  & $0.0$ & 98.9 & 0.3 & 0.1 & 0.0 & 0.7 \\
& & $0.7$ & 94.9 & 0.7 & 1.0 & 0.7 & 2.8 \\
& & $0.8$ & 92.9 & 1.0 & 0.7 & 1.3 & 4.1 \\
& & $1.0$ & 85.4 & 2.0 & 1.2 & 2.7 & 8.7 \\
\midrule
\multirow{11}{*}{{LLaDA}} 
& \multirow{4}{*}{Sentiment (\Cref{tab:combined_results_final})} 
  & $0.0, 0.0$ & 62.2 & 6.0 & 4.1 & 6.4 & 21.2 \\
& & $0.7, 0.75$ & 1.0 & 0.6 & 1.6 & 5.0 & 91.9 \\
& & $0.8, 0.85$ & 0.2 & 0.4 & 1.0 & 4.2 & 94.1 \\
& & $0.95, 1.0$ & 0.0 & 0.0 & 0.1 & 2.4 & 97.4 \\
\cmidrule{2-8}
& \multirow{4}{*}{Toxicity (\Cref{tab:combined_results_final})} 
  & $0.0, 0.0$ & 99.9 & 0.0 & 0.0 & 0.0 & 0.1 \\
& & $0.7, 0.75$ & 53.1 & 1.0 & 1.1 & 1.4 & 43.3 \\
& & $0.8, 0.85$ & 37.7 & 1.1 & 0.8 & 1.3 & 59.1 \\
& & $0.95, 1.0$ & 27.4 & 1.1 & 0.7 & 1.6 & 69.2 \\
\cmidrule{2-8}
& \multirow{3}{*}{Detoxification (\Cref{tab:detoxification})} 
  & $0.0, 0.0$ & 58.3 & 0.6 & 0.4 & 0.6 & 40.1 \\
& & $0.7, 0.75$ & 89.0 & 0.2 & 0.2 & 0.3 & 10.4 \\
& & $0.8, 0.85$ & 91.8 & 0.1 & 0.1 & 0.2 & 7.8 \\
\bottomrule
\end{tabular}
\caption{Distribution of ILRR generations (\%) across classifier score ranges at different steering scales $\alpha$.}
\label{tab:score_distributions}
\end{table*}

\section{Analysis and Ablation}
We analyze the impact of several key parameters
on steering effectiveness, while all remaining parameters are kept fixed. Perplexity (PPL) is evaluated using GPT2-XL \citep{radford2019language}.
\label{app:analysis_and_ablation}

\subsection{Steering Layers $L_s$}
\label{app:layers_ablation}
We analyze the effect of steering depth on negative sentiment control on LLaDA, using single and two-layer interventions (\Cref{fig:layer_single,fig:layer_pairs}). We observe that steering effectiveness peaks at mid-to-late layers, while earlier layers are consistently less effective. Furthermore, intervening at two layers generally improves performance compared to single-layer steering, particularly when both layers are selected from the mid-to-late range.

To further understand the underlying geometry of the latent representations and the semantic meaning captured by the average pooling operation, we visualize the latent space at different  layers using Principal Component Analysis (PCA). By projecting the sequence average-pooled representations of positive and negative sentiment reference texts (\Cref{fig:pca_layer_6,fig:pca_layer_20}), we observe that early layers (e.g., 6) exhibit entangled representations with no clear separation. In contrast, mid-to-late layers (e.g., 20) appear to form distinct, linearly separable semantic subspaces. This suggests that later layers are more capable of interpreting high-level semantic features and providing a cleanly structured space, which directly aligns with our results, demonstrating that these deeper layers are significantly more effective for steering.

Furthermore, by tracking the generation's sequence average-pooled representations across the denoising timesteps at layer 25 using a positive sentiment reference for steering, we can observe the direct mechanics of our method. As shown in \Cref{fig:trajectory_layer_25}, while the unsteered baseline's trajectory naturally drifts away from the target attribute, the ILRR intervention successfully redirects the generation's latent trajectory, guiding it directly into the target positive semantic subspace defined by the reference.

\subsection{Tunability}
\label{app:tunability}
\textbf{Steering Scale $\alpha$.} We examine the relationship between steering scale and steering intensity on positive sentiment control on LLaDA (\Cref{fig:steering_scale}). We additionally report 4-gram overlap values, which measure the percentage of 4-token sequences in the generation that identically match the reference \citep{papineni-etal-2002-bleu}. We find that while steering accuracy improves significantly with $\alpha$, reaching nearly $100\%$ at $\alpha \ge 0.8$, the 4-gram overlap remains low, indicating that the improved steering is not driven by verbatim reuse of reference text, but rather by semantic alignment. \\
\textbf{Steering Steps $T_s$.} We analyze the impact of limiting the steering intervention to specific stages of the denoising process (\Cref{fig:steering_steps}), on positive sentiment control on LLaDA. We observe that effective steering is achievable across all settings, though the early stage proves significantly stronger than the late stage, which yields the lowest accuracy. This suggests that early denoising steps offer greater leverage for shaping high-level semantics, while later stages are less amenable to coarse semantic control.

\subsection{Reference Quality}
\label{app:reference_quality_ablation}
To better understand ILRR's sensitivity to the grammatical structure and semantic integrity of the reference sequence, we conduct a graded degradation analysis on the positive sentiment task on LLaDA. We evaluate generation fluency (Flu.) utilizing GPT-4o \citep{openai2024gpt4ocard} as an LLM-as-a-judge (1=Poor, 5=Excellent). As shown in \Cref{tab:ref_quality}, we analyze three axes of degradation: (i) \textbf{Data Corruption}. We degrade the reference by replacing varying ratios of it with $"."$ tokens (sparsity) or random tokens (noise). We observe that while steering accuracy remains relatively robust to minor data loss ($87\%$ accuracy at $25\%$ sparsity), generation fluency noticeably degrades. The method's effectiveness drops across all metrics when critical signal density falls further or when significant noise is introduced. \textbf{(ii) Structural Shuffling.} We divide the reference text into chunks of varying lengths and randomly shuffle them. Preserving larger phrases ($\sim$10-word chunks) maintains high steering accuracy and fluency. However, as shuffling becomes overly granular (2-word chunks, or full word-level shuffling), underlying semantics are disrupted and the steering degrades. This indicates that ILRR steers via contextualized representations rather than acting as a simple ``bag-of-words'' guide.
\textbf{(iii) Semantic Quality.} We test three semantic edge cases: (a) a midway shift from positive to negative sentiment, (b) a highly repetitive reference echoing the exact same idea, and (c) grammatically correct gibberish with no clear meaning. A mid-sequence semantic shift significantly drops target accuracy, demonstrating that a target attribute must be globally consistent throughout the reference, as shifts naturally cascade into the generated text. Gibberish lacks a distinct target subspace, resulting in poor steering accuracy. Conversely, a repetitive reference achieves high accuracy but induces mode collapse, severely destroying lexical diversity. 

Overall, the results suggest that an effective reference requires semantic coherence and fluency, global consistency, and lexical diversity.

\subsection{Kernel Size $k$}
\label{app:kernel_ablation}
We analyze the effect of the pooling window size on toxicity control on LLaDA (\Cref{fig:kernels}). We observe that small kernels ($k \le 4$) yield significantly high 4-gram overlap and a high PPL, indicating that the steering becomes overly localized. Because each signal aggregates activations from only a small number of tokens, it exhibits higher variance and increased sensitivity to fine-grained token-level fluctuations. This suggests that the model is driven toward overly aggressive local alignment with the reference without preserving global semantic or syntactic structure, resulting in increased copying behavior and degraded likelihood. As the kernel size increases to the mid-range ($k > 4$), we identify a sweet spot. We maintain strong steering accuracy while the 4-gram copying and PPL drop significantly. This confirms that coarser pooling effectively abstracts high-level attributes, allowing for robust steering without degrading into simple replication.

\subsection{Spatially Modulated Steering}
\label{app:spatially_modulated_steering_ablation}
To evaluate the robustness of our spatially modulated steering mechanism (\Cref{sec:longer_seq}), we analyze its sensitivity to varying modulation wave frequencies and waveform shapes on positive sentiment control on MDLM. \\
\textbf{Waveform shape.}
We analyze the sensitivity to the choice of modulation waveform. As shown in \Cref{tab:waveforms}, the results reveal three categories of behavior:
    (i) Broad alternating waveforms (cosine, linear ramp, square, concave parabola) achieve a strong balance of accuracy and PPL. This indicates that broad and alternating signal coverage is more critical than the exact mathematical shape. The concave parabola achieves the highest accuracy within this group but incurs a PPL degradation, likely due to its aggressive peak intensity.
    (ii) Uniform scale (no modulation) achieves the highest absolute accuracy but significantly degrades PPL. This isolates the alternating intensity as the critical mechanism balancing control and quality.
    (iii) Monotonic and pulse-like forms (sigmoid, exponential variants, Gaussian) yield significantly poorer steering performance. These functions concentrate the steering signal too narrowly, leaving large portions of the sequence unguided, failing to provide the sustained semantic signal necessary for effective guidance.
These results indicate that our method is not strictly tied to the cosine function. Rather, cosine serves as a reliable representative within a broader class of waveforms that distribute the steering signal across the sequence through smoothly varying or alternating intensity. \\
\textbf{Wave Frequency.} We examine the impact of the modulation wave frequency $f$ on generation PPL and control (\Cref{fig:modulated_steering}), using a cosine waveform. The results indicate that the method is not highly sensitive to the specific choice of wave frequency, with performance remaining relatively stable across the different configurations. This suggests that the benefits arise from the modulation pattern itself rather than careful tuning of $f$.

\section{Qualitative Examples}
\label{app:qualitative_examples}
\begin{table*}
\label{tab:qualitative_examples}
\begin{center}
\begin{small}
\begin{tabularx}{\linewidth}{p{0.10\linewidth} p{0.10\linewidth} X X}
\toprule
\textsc{Attribute} & \textsc{Prompt} & \textsc{Reference Text} & \textsc{Generated Output} \\
\midrule

\textbf{Positive \newline Sentiment} & 
The painting &
is incredibly detailed and skillful, standing as a true testament to the artist's immense talent and their ability to bring joy through visual art, a stunning display of creativity that enriches the lives of everyone who sees it and serves as a constant reminder of the beauty that human hands are capable of creating with passion. & 
was a reflection of the beauty and complexity of the landscape, and the artist's ability to capture the essence of the natural world. It was a stunning work of art that captured the spirit of the world, and the people who saw it. \\
\addlinespace[0.7em]

\textbf{Negative \newline Sentiment} & 
The movie &
was relentlessly depressing and bleak, lacking any hope or joy and leaving me feeling empty, sad, and hollow inside, a cinematic experience that felt more like an emotional punishment than entertainment, dragging the viewer down into a pit of despair and leaving them there without a ladder to climb back out to the light. & 
is about the darkness of the human soul, and it portrays a sense of emptiness, loneliness, and despair, but it does so in a way that feels like a human experience. However, at the end of the movie, the characters are left with \\
\addlinespace[0.7em]

\textbf{Toxicity} & 
The last time &
you tried to help, you ruined everything because you are incompetent, clumsy, and utterly useless at everything you do, turning a simple task into a disaster because you are incapable of doing anything right, proving once again that you are a burden on everyone around you who has to clean up your mess. & 
I had to do that, I was so stupid, clumsy, and totally incompetent, and I failed at everything. It turned into a complete disaster for me, I was incapable of doing anything, and I was nothing but a burden to my family. \\
\addlinespace[0.7em]

\textbf{Science} & 
The central theme &
Rigorous scientific inquiry demands a strict adherence to the established scientific method, where trained researchers objectively observe natural processes, formulate highly testable hypotheses, and execute carefully controlled laboratory experiments to eliminate confounding variables, thereby producing highly reliable, reproducible experimental data. & 
of this story is the pursuit of truth and the importance of scientific methodology. The author of the story emphasizes the importance of observation, hypotheses, and the design of experiments. In science, confounding variables must produce reliable, reproducible data that can be used to \\
\addlinespace[0.7em]

\textbf{Religion} & 
The relationship &
Throughout all of recorded human history, established religious traditions have played an absolutely central role in shaping human culture, heavily emphasizing the paramount importance of spiritual devotion, divine revelation, and the ultimate pursuit of heavenly salvation through regular daily prayer, strict adherence to sacred doctrines, and active participation in holy rituals. & 
between the human and the divine is complex, religious traditions have played a role in the human understanding of the world, including the belief in the salvation of the soul, the pursuit of enlightenment and happiness, and the practice of prayer and worship. \\
\addlinespace[0.7em]

\textbf{Space} & 
The key aspect &
Astronomical observations through powerful modern telescopes reveal a breathtaking cosmos where massive solar systems orbit around supermassive black holes, and interstellar dust clouds give birth to entirely new celestial bodies, highlighting the incredible astronomical complexity of planetary atmospheres, lunar phases, and the immense gravitational forces. &
of the problem is the formation of the universe, which is a complex system of galaxies, stars, black holes, and gas clouds that led to the formation of galaxies. The goal is to understand the computational physics of the formation of the universe. \\
\addlinespace[0.7em]

\bottomrule
\end{tabularx}

\end{small}
\end{center}
\caption{Qualitative examples of text generations with ILRR. We show the prompt, the reference used to steer, and the resulting generation. Generations are fixed to 50 tokens, while references are truncated to align with this length. Note: The toxicity example contains potentially offensive language.}
\end{table*}

\end{document}